\documentclass{article}

\PassOptionsToPackage{numbers}{natbib}
\usepackage[preprint]{neurips_2025}

\usepackage[utf8]{inputenc} 
\usepackage[T1]{fontenc}    
\usepackage{hyperref}       
\usepackage{url}            
\usepackage{booktabs}       
\usepackage{amsfonts}       
\usepackage{nicefrac}       
\usepackage{microtype}      

\usepackage{caption}

\usepackage{graphicx} 
\usepackage{colortbl}
\usepackage{color}
\usepackage{pifont}
\usepackage{wrapfig}
\usepackage[font=footnotesize]{caption}
\usepackage{wrapfig}
\usepackage{placeins}

\usepackage{amssymb}
\usepackage{amssymb}

\usepackage{times}
\usepackage{latexsym}
\usepackage[dvipsnames]{xcolor}

\usepackage[T1]{fontenc}
\usepackage[utf8]{inputenc}

\usepackage{microtype}

\usepackage{inconsolata}

\usepackage{graphicx}
\usepackage{booktabs}
\usepackage{multirow}
\usepackage{makecell}
\usepackage{geometry}
\usepackage{tcolorbox}
\usepackage{array}
\usepackage{colortbl}
\usepackage{graphicx}
\usepackage{CJK}

\usepackage{lscape} 
\usepackage{subfig} 
\usepackage{caption} 
\usepackage{pifont} 
\newcommand{\cm}{\ding{51}}

\usepackage{tcolorbox}
\tcbuselibrary{theorems}

\definecolor{lightblueshade}{rgb}{0.8,0.9,1}
\definecolor{bluex}{rgb}{0.27, 0.42, 0.81}
\definecolor{purplex}{HTML}{9564bf}
\definecolor{red3}{HTML}{C52A20}
\definecolor{red2}{HTML}{B36A6F}
\definecolor{red1}{HTML}{FFb5b5}
\definecolor{purple}{HTML}{B36A6F}
\definecolor{darkyellow}{HTML}{D5BA82}
\definecolor{blue1}{HTML}{A0C0E0}
\definecolor{blue2}{HTML}{C4E4E3}
\definecolor{green1}{HTML}{A1D0C7}
\definecolor{green2}{HTML}{BFF6BA}
\definecolor{green3}{HTML}{028100}
\definecolor{teal}{HTML}{508AB2}
\definecolor{purple1}{HTML}{8d3a94}
\definecolor{olivegreen}{rgb}{0.33, 0.42, 0.18}

\newtcbtheorem[number within=section]{exmp}{Example}%
{colback=green2!5,colframe=blue1,fonttitle=\bfseries, left=.02in, right=.02in,bottom=.02in, top=.02in}{exmp}

\title{MM-IQ: Benchmarking Human-Like Abstraction and Reasoning in Multimodal Models}

\author{Huanqia Cai$\thanks{caihuanqia19@mails.ucas.ac.cn}$ \quad
Yijun Yang \quad 
Winston Hu  
 \\~\\
Tencent Hunyuan Team\\
}

\begin{document}

\maketitle

\begin{figure}[h]
  \vspace{-1.5em}
    \centering
    \includegraphics[width=\textwidth]{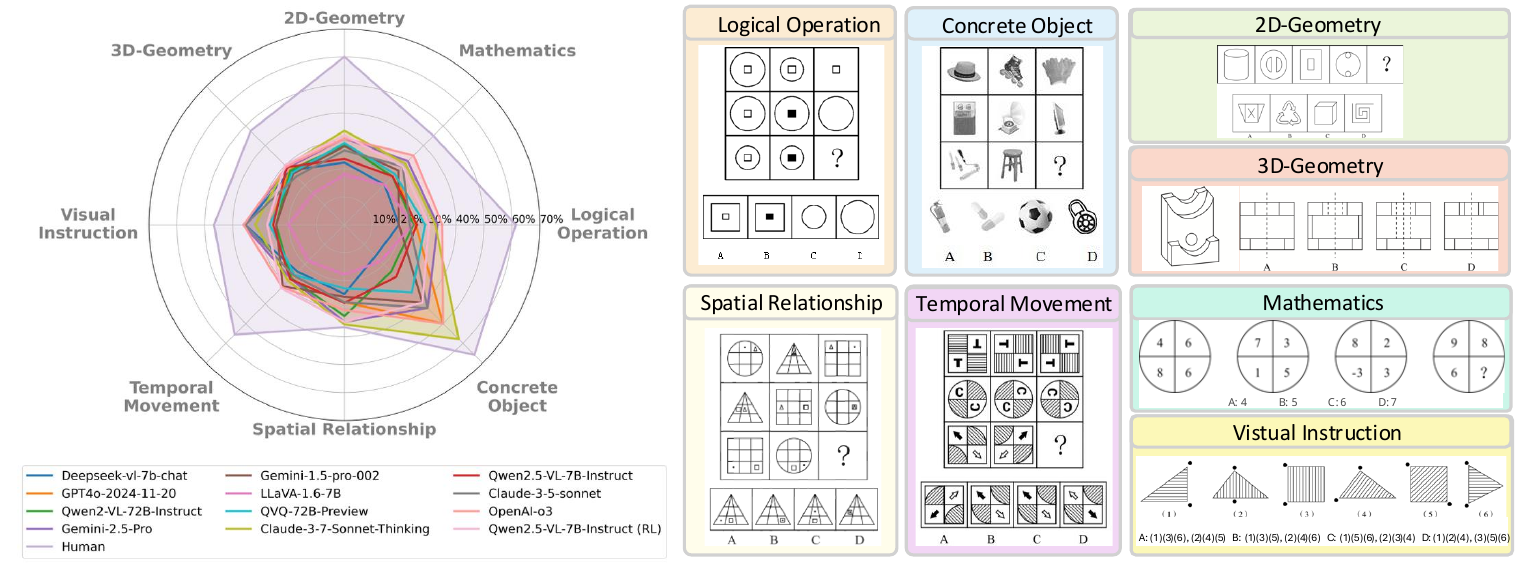}
    \caption{\textbf{Left}: Accuracy of large multimodal models vs. humans across eight reasoning paradigms of MM-IQ. \textbf{Right}: Visual examples of MM-IQ's reasoning paradigms (Detailed information can be found in Section~\ref{sec:reasoning pattern}).\looseness-1}
    \label{fig:MLLMs' performance}
\end{figure}

\begin{abstract}
IQ testing has served as a foundational methodology for evaluating human cognitive capabilities, deliberately decoupling assessment from linguistic background, language proficiency, or domain-specific knowledge to isolate core competencies in abstraction and reasoning. Yet, artificial intelligence research currently lacks systematic benchmarks to quantify these critical cognitive capabilities in multimodal systems. To address this crucial gap, we propose \textbf{MM-IQ}, a comprehensive evaluation framework, which comprises a large-scale training set with 4,776 visual reasoning problems and 2,710 meticulously curated test items spanning 8 distinct reasoning paradigms.
Through systematic evaluation of existing open-source and proprietary multimodal models, our benchmark reveals striking limitations: even state-of-the-art architectures achieve only marginally superior performance to random chance (33.17\% vs. 25\% baseline accuracy). This substantial performance chasm highlights the inadequacy of current multimodal models in approximating fundamental human reasoning capacities, underscoring the need for paradigm-shifting advancements to bridge this cognitive divide. Moreover, inspired by the recent surge of large reasoning models, we also release a multimodal reasoning model as the baseline that is trained via reinforcement learning with verifiable reward functions, reaching competitive performance to the state-of-the-art with a notably smaller model size. 

\raisebox{-0.3\height}{\hspace{0.1cm}\includegraphics[width=0.41cm]{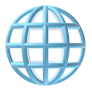}} \small \textbf{\mbox{Homepage:}} \href{https://acechq.github.io/MMIQ-benchmark/}{acechq.github.io/MMIQ-benchmark/}


\end{abstract}


\section{Introduction}
The rapid advancement of large multimodal models (LMMs) has intensified debates about their capacity for human-like abstraction and reasoning. While existing benchmarks evaluate specialized capabilities such as OCR, object localization, and medical image analysis~\cite{liu2023hidden,yue2024mmmu,liu2025mmbench}, these task-specific metrics fail to quantify the critical cognitive dimensions in multimodal systems. This limitation mirrors a long-standing challenge in human cognitive assessment: early methods conflated domain knowledge with innate reasoning ability until IQ testing emerged to isolate core cognitive competencies through language- and knowledge-agnostic evaluations~\cite{snow1984topography}. Inspired by this paradigm, we argue that multimodal intelligence evaluation should also similarly decouple linguistic proficiency and task-specific knowledge from the measurement of abstract reasoning capacities.

Abstract Visual Reasoning (AVR) offers a plausible solution to the above challenge. As shown in Figure~\ref{fig:Logical operation reasoning}, AVR problems usually contain visual puzzles with simple 2D/3D shapes. Solving these problems requires identifying and understanding the underlying abstract rules and generalizing them to novel configurations. 
Although there exists a wide range of AVR benchmarks, e.g., RAVEN~\cite{zhang2019raven}, Bongard-LOGO~\cite{nie2020bongard}, and SVRT~\cite{fleuret2011comparing}, most of them have limited input modalities, reasoning paradigms, and restricted problem configurations, which can lead to biased evaluation results~\cite{van2021much}.

To this end, we propose MM-IQ, a comprehensive AVR benchmark comprising 2,710 meticulously curated test items spanning 8 distinct reasoning paradigms. Like human IQ tests, MM-IQ fully eliminates domain-specific and linguistic biases while systematically diversifying problem configurations to prevent pattern memorization, presenting striking challenges for LMMs: even state-of-the-art models achieve only 33.17\% accuracy, marginally exceeding random chance (25\%) but far below human-level performance (51.27\%). This substantial performance chasm highlights the inadequacy of current LMMs in approximating fundamental human reasoning capacities, underscoring the need for paradigm-shifting advancements to bridge this cognitive divide.
By applying IQ-testing principles to multimodal models, MM-IQ fills a critical gap in existing multimodal benchmarks, e.g., MMBench~\cite{liu2025mmbench} and MMMU~\cite{yue2024mmmu} that focus on broad task coverage rather than core reasoning abilities. Our results demonstrate that current architectures lack the intrinsic abstraction abilities necessary for human-like intelligence, shedding light on potential directions toward developing systems capable of genuine cognitive adaptation.\looseness-1

To facilitate further research and support the community in building and refining models with stronger abstract reasoning abilities, we purposely release a high-quality training set consisting of 4,776 question-answer pairs. Our preliminary experiments show that leveraging reinforcement learning (RL) on this training set can lead to notable performance improvement, suggesting that accurate, high-quality data and appropriately chosen training algorithms may help close the reasoning gap between current LMMs and humans.

\section{Related Work} 
\label{sec:Categories}
Following ~\cite{malkinski2023review,jiang2024marvel, malkinski2022deep}, all existing AVR benchmarks, including our MM-IQ, can be cataloged along three dimensions: input shape, problem configuration, and reasoning paradigm, as shown in Table ~\ref{tab:dataset}.
Input shape refers to the input forms of the objects in the given image, which contributes to evaluating models' cognition abilities of different shapes. Diverse problem configurations assess models' abstract reasoning capabilities across multi-dimensional aspects, including pattern recognition (Raven’s Progressive Matrices~\cite{raven2003raven}), analogical transfer ability (Visual Analogy~\cite{hill2019learning}), discrimination ability (Odd-one-out~\cite{mandziuk2019deepiq}), extrapolation and generalization ability (Visual Extrapolation~\cite{webb2020learning}), and numerical reasoning ability (Arithmetic Reasoning~\cite{zhang2020machine}), etc. MM-IQ's inclusion of diverse problem configurations ensures a thorough evaluation of multimodal models' abstract reasoning capabilities across various AVR problems. Reasoning paradigm is a more fine-grained category that evaluates LMMs' abstract reasoning capabilities, like logical deduction, temporal and spatial cognition, geometric, etc. It includes various reasoning paradigms such as temporal movement, spatial relationships, logical operations, and both 2D and 3D geometry, which are based on the internal forms, relationships, and numbers of objects in the given image. Existing benchmarks have only three paradigms on average except for MARVEL, which has five ones, but its quantity is relatively small. Although RAVEN~\citep{zhang2019raven},  G-set~\citep{mandziuk2019deepiq}, VAP~\citep{hill2019learning}, and DOPT~\citep{webb2020learning} have more than 1,000 instances, all of their data are generated by computer programs, which lack diversity and complexity~\cite{chollet2019measure}. MM-IQ test set comprises a total of 2,710 meticulously selected problems, 3x larger than MARVEL, and covers a diverse spectrum of 8 fine-grained reasoning paradigms.

\textbf{Training Reasoning Models with RL.}
RL has recently emerged as a pivotal approach for incentivizing the reasoning capabilities of both large language models (LLMs) and large multimodal models (LMMs). Several representative studies have demonstrated that RL-based training strategies can significantly improve models' performance on complex reasoning tasks. For instance, DeepSeek-R1~\cite{guo2025deepseek} combines Group Relative Policy Optimization (GRPO)~\cite{shao2024deepseekmath} and a verifiable, rule-based reward method to achieve substantial improvements in math and coding benchmarks. Kimi-1.5~\cite{team2025kimi} also observes a similar phenomenon that pure RL can train LLMs to solve complex reasoning tasks via long CoT context characterized by emergent reflection, backtracking search, and self-verification. In the multimodal domain, models such as Vision-R1~\cite{huang2025vision} and R1-V~\cite{chen2025sft} have applied R1-style RL to specific downstream tasks, including geometric reasoning and object counting, exhibiting promising progress in reproducing RL's success for LMMs. These advancements underscore the growing importance of RL techniques in pushing the limit of abstract reasoning for unimodal and multimodal AI systems and highlight the necessity of comprehensive benchmarks like MM-IQ to evaluate and foster such progress.


\begin{table*}[t]
\centering
\resizebox{\textwidth}{!}{%
\begin{tabular}{ll|c|c|c|c|c|c|c|c|c|c}
\toprule
\multicolumn{2}{c|}{}                                            & RAVEN$\ast$ & G-set$\ast$ & VAP$\ast$  & SVRT &  DOPT$\ast$ & ARC & MNS & IQTest & MARVEL & \textbf{MM-IQ} \\ \midrule
\multicolumn{1}{l}{\multirow{3}{*}{\textbf{Input Shape}}}   & Geometric            & \cm  & \cm   & \cm &                                  &\cm  & \cm  &\cm & \cm & \cm   & \cm  \\

\multicolumn{1}{l}{}                               & Abstract             &      &       &                                & \cm  &     &   &  & \cm  & \cm  & \cm    \\

\multicolumn{1}{l}{}                               & Concrete Object             &      &       &                                       &   &     &    &  & & & \cm     \\ 

\midrule

\multicolumn{1}{l}{\multirow{6}{*}{\textbf{Problem Configuration}}} & Raven’s Progressive Matrices~\cite{raven2003raven}             &  \cm    &\cm       &                                  &      &  &   &  & \cm &  \cm & \cm \\

\multicolumn{1}{l}{}                               & Visual Analogy~\cite{hill2019learning}               &      &     &   \cm                                     &      &     &   &  &  & \cm &\cm    \\

\multicolumn{1}{l}{}                               & Odd-one-out~\cite{mandziuk2019deepiq}               &  & \cm &                                        &      &     &   & & \cm & & \cm   \\

\multicolumn{1}{l}{}                               & Visual Extrapolation~\cite{webb2020learning}                 &      &       &                                     &    & \cm    &   & & \cm &\cm  & \cm    \\

\multicolumn{1}{l}{}                               & Arithmetic Reasoning~\cite{zhang2020machine}            &      &       &                                             &      &     &   &  \cm & \cm &  & \cm   \\ 

\multicolumn{1}{l}{}                               & Visual Grouping               &      &       &                                          & \cm     &     &   &  &  &  & \cm   \\ 

\midrule

\multicolumn{1}{l}{\multirow{9}{*}{\textbf{Reasoning Paradigm}}}       & Temporal Movement    & \cm  & \cm   &                                &      & \cm & \cm  &  & & \cm   & \cm  \\
\multicolumn{1}{l}{}                               & Spatial Relationship &      &       &                                        & \cm   &     & \cm   & & & \cm   & \cm  \\
\multicolumn{1}{l}{}                               & 2D-Geometry          &      &       &                               & \cm  &     &    &  &\cm & \cm   & \cm \\
\multicolumn{1}{l}{}                               & 3D-Geometry          &      &       &                                       &      &     & &  &\cm  & \cm  & \cm \\ 

\multicolumn{1}{l}{}                               & Logical Operation          &  \cm    &       & \cm                                 &      &     &   &  &  &  & \cm \\
\multicolumn{1}{l}{}                               & Concrete Object          &      &       &                                         &      &     &   &  &  &  & \cm \\
\multicolumn{1}{l}{}                               & Visual Instruction          &      &       &                                         &      &     &   &  &  &  & \cm \\
\multicolumn{1}{l}{}                               & mathematics           & \cm     &  \cm     &  \cm                                       & \cm     & \cm    & \cm  & \cm & \cm & \cm & \cm \\

\bottomrule
\multicolumn{2}{c|}{Dataset Size}                  & 14,000  & 1,500      &   100,000   &     23                         &  95,200    & 600     &  - & 228 & 770 & \textbf{2,710 + 4,776} \\ \bottomrule
\end{tabular}%

}
\caption{Comparison between our \textsc{MM-IQ} and related benchmarks: RAVEN$\ast$~\citep{zhang2019raven}, G-set$\ast$~\citep{mandziuk2019deepiq}, VAP$\ast$~\citep{hill2019learning}, SVRT~\citep{fleuret2011comparing}, DOPT$\ast$~\citep{webb2020learning},
ARC~\citep{chollet2019measure}, MNS~\cite{zhang2020machine},
IQTest~\cite{lu2023mathvista},
MARVEL~\cite{jiang2024marvel}. $\ast$ denotes that the dataset is automatically produced through procedural content generation.\looseness-1}
\label{tab:dataset}
\vspace{-0.3cm}
\end{table*}

\section{Construction of MM-IQ}

Three features distinguish MM-IQ from other existing benchmarks for LMMs: (1) MM-IQ adopts data from professional and authoritative examinations and performs rigorous quality control, which ensures its correctness and validity; (2) MM-IQ is a comprehensive AVR benchmark for evaluating the intelligence of LMMs, comprising a total of 2,710 problems and covering a diverse spectrum of 8 fine-grained reasoning paradigms; (3) MM-IQ also provides a high-quality training set containing 4,776 carefully selected problem-answer pairs to accelerate the future research toward an excellent multimodal reasoner like humans.


\subsection{Data Collection}
\label{sec:data collection}
The collection of MM-IQ involves three stages. Initially, we examined existing AVR datasets~\cite{zhang2019raven, mandziuk2019deepiq, chollet2019measure, nie2020bongard} and discovered that most of them are generated by hand-coded procedures. Although programmatic synthesis can produce substantial amounts of data, it often lacks the necessary diversity. Hence, we chose to collect AVR problems from existing resources. Following ~\cite{liu2020logiqa, jiang2024marvel, zhang2024cmmmu}, we collected problems from publicly available questions of the National Civil Servants Examination of China. These problems are specifically designed to evaluate civil servant candidates’ critical thinking and problem-solving skills, and they meet our criteria for both quantity and diversity. The collected data underwent a rigorous filtering process conducted by human annotators to eliminate any low-quality entries. The filtering principle is that the problems can be solved only by extracting and utilizing high-level abstract reasoning information based on visual inputs.

To create a systematic and comprehensive benchmark, we proceeded to categorize the data into different reasoning paradigms and further augmented underrepresented paradigms. Based on the descriptions of collected problems, we classified them into the corresponding reasoning paradigms. Additionally, we identified the common attributes of each paradigm's problems, such as attributes and entity types, and supplemented those with fewer instances to ensure that each fine-grained attribute or entity type had sufficient problems.

The final stage involved a more thorough cleaning of the collected data through deduplication and extraction of the final answers. We performed deduplication in two ways. The first way was to employ the MD5 hashing algorithm to find the same images and remove them if their input text was the same. Secondly, we utilized the problems' corresponding information, where similar ones were considered suspected duplicates, and then reviewed by human annotators based on the input image and corresponding information to identify and eliminate duplications. 

Additionally, the final answers were extracted by human annotators to facilitate efficient evaluation later. To further support the development of the open-source community, we also translated all content of questions and answers from Chinese to English based on GPT-4, resulting in a bilingual version of the dataset. All translations were verified by humans to ensure their correctness. Specifically, the data distribution of the reasoning paradigms is shown in the Figure~\ref{fig:error_distribution_and_Data_distribution_pile}, where concrete object and visual instruction are less than 2\% since they are rare in the existing data.

Based on the above data collection process, we constructed a comprehensive and well-curated evaluation set, which contains a total of 2,710 samples covering eight reasoning paradigms. Following a similar processing pipeline, we further constructed a training set. To mitigate the risk of test set leakage, we utilized the image encoder of CLIP-ViT-B32~\cite{radford2021learning} to extract image features for the training set and removed any images whose cosine similarity with those in the test set exceeded 97.5\%. As a result, the training set comprises 4,776 samples.

\subsection{Reasoning Paradigms of MM-IQ}
\label{sec:reasoning pattern}

For simplicity and consistency, we follow MARVEL~\cite{jiang2024marvel}, a dataset evaluating LMMs' AVR ability but 3x smaller than ours, and extend its taxonomy to 8 categories, including logical operation, mathematics, 2D-geometry, 3D-geometry, visual instruction, temporal movement, spatial relationship, and concrete object. Notably, we merge mathematical and quantity categories from MARVEL's taxonomy into mathematics to align more closely with our taxonomy.

\begin{figure*}[h!]
    \centering
    \begingroup
    \setlength{\fboxsep}{2pt}
    \setlength{\fboxrule}{1pt}
    \small 
    \begin{tcolorbox}[colframe=RoyalBlue, colback=white, boxrule=0.5pt, left=1mm, right=1mm, top=1mm, bottom=1mm]
    \begin{minipage}[t]{0.45\textwidth}
        \textbf{Prompt:} \small Choose the most appropriate option from the given four choices to fill in the question mark, so that it presents a certain regularity:\\[1pt]
        \centering
        \includegraphics[width=0.6\linewidth]{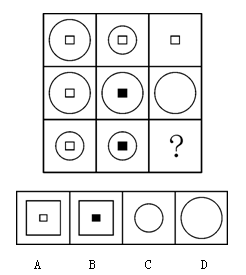}
    \end{minipage}
    \hfill
    \begin{minipage}[t]{0.5\textwidth}
        \textbf{LMM's Response:} 
        \begin{tcolorbox}[colframe=RoyalBlue, colback=gray!10, boxrule=1pt, left=1mm, right=1mm, top=1mm, bottom=1mm]
        The first two panels in each row have similar parts to the third panel.\par

        In the first and second rows, the third panel is obtained by removing the differences and finding the similarities between the first two panels; the same rule is applied to the third row, and the small circle is obtained by removing the differences and finding the similarities between the first two panels. Only option C meets the requirements. \par

        Therefore, option C is selected.
        \end{tcolorbox}
        \textbf{Ground Truth:} \textbf{C}
    \end{minipage}
    \end{tcolorbox}
    \endgroup
    \caption{A visualized example of logical operation paradigm.}
    \label{fig:Logical operation reasoning}
\end{figure*}

\textbf{Logical Operation} refers to the application of logical operators, such as AND (conjunction), OR (disjunction), XOR (exclusive disjunction), etc. This reasoning process involves observing and summarizing the abstract logical operations represented in the given graphics to derive general logical rules, which can then be applied to identify the required graphics. An example of reasoning involving the AND operation is shown in Figure~\ref{fig:Logical operation reasoning}.

\textbf{2D-Geometry} encompasses two distinct categories. The first category involves understanding the attribute patterns of the provided 2D geometric graphics, such as symmetry, straightness, openness, and closure, and making analogies or extrapolations based on these attributes. The second category focuses on graphic splicing, which entails identifying a complete pattern that can be formed by assembling existing 2D geometric fragments. Together, these two types assess the capability of LMMs to perceive geometric shapes from both local and global perspectives. A visualized example of 2D-geometry reasoning concerning the symmetry property is shown in Figure~\ref{fig:2D geometry reasoning} (see Appendix~\ref{appendix:example of reasoning paradigms} for details).

\textbf{3D-Geometry} can be categorized into three categories. The first category assesses the capability of LMMs to perceive 3D geometry comprehensively by observing a polyhedron and identifying the required view from a specific direction. The second category is analogous to 2D graphic splicing, but it involves basic fragments and target objects that are three-dimensional in nature. The third category evaluates LMMs' comprehension of the interior structure of a 3D solid shape with the goal of identifying a cross-sectional view of the solid. An example of 3D-geometry reasoning for the specific directional view is shown in Figure~\ref{fig:3D geometry reasoning} (see Appendix~\ref{appendix:example of reasoning paradigms} for details).

\textbf{Visual Instruction} employs visual cues such as points, lines, and arrows to highlight key areas necessary for solving visual puzzles. Unlike other reasoning paradigms, this approach allows test-takers to concentrate solely on these visual indicators rather than requiring a comprehensive observation of the entire panel. A visualized example of visual instruction reasoning with arrows is shown in Figure~\ref{fig:visual instruction reasoning} (see Appendix~\ref{appendix:example of reasoning paradigms} for details).

\textbf{Temporal Movement} focuses on changes in position or movement, including translation, rotation, and flipping. This paradigm encompasses several problem configurations discussed in Section~\ref{sec:Categories}, including Raven’s Progressive Matrices, Visual Analogy, and Visual Extrapolation. A visualized example of temporal movement reasoning involving rotation is shown in Figure~\ref{fig:temporal movement reasoning} (see Appendix~\ref{appendix:example of reasoning paradigms} for details).

\textbf{Spatial Relationship} examines the static relative positional relationships among objects. This paradigm also encompasses various problem configurations, including Raven’s Progressive Matrices, Visual Analogy, Visual Extrapolation and Visual Grouping. An example of spatial relationship reasoning is shown in Figure~\ref{fig:spatial relationship reasoning} (see Appendix~\ref{appendix:example of reasoning paradigms} for details).

\textbf{Concrete Object} involves real-world objects, such as vases, leaves, or animals, and requires LMMs to categorize these objects based on their characteristics, which may require external knowledge to solve. A visualized example of concrete object reasoning is shown in Figure~\ref{fig:concrete object reasoning} (see Appendix~\ref{appendix:example of reasoning paradigms} for details).

\textbf{Mathematics} evaluates LMMs' ability to reason about quantity, numbers, and arithmetic operations through visual inputs. This paradigm contains two types of tasks. The first type involves perceiving basic graphical elements, such as points, angles, lines, and planes, and applying arithmetic operations to these elements. The second type involves identifying an arithmetic expression that is satisfied by the numbers in the given images and determining the missing number based on the four fundamental operators: addition, subtraction, multiplication, and division. Examples of the two types of mathematics reasoning are shown in Figure~\ref{fig:quantity reasoning} and Figure~\ref{fig:arithmetic visual reasoning}. The intersection point is the basic element used in Figure~\ref{fig:quantity reasoning} (see Appendix~\ref{appendix:example of reasoning paradigms} for details). \looseness-1




\begin{table*}[t!]
    \centering
    \caption{\textbf{LMMs and Human Performance on MM-IQ (\%)}. Abbreviations adopted: \textbf{LO} for Logical Operation; \textbf{2D-G} for 2D-Geometry; \textbf{3D-G} for 3D-Geometry; \textbf{VI} for Visual Instruction; \textbf{TM} for Temporal Movement; \textbf{SR} for Spatial Relationship; \textbf{CO} for Concrete Object. Qwen2.5-VL-7B-Instruct (RL) denotes our RL-trained baseline. }
    \label{tab:performance}
    \resizebox{\textwidth}{!}{ 
        \begin{tabular}{lccccccccc}
            \toprule
            \rowcolor{gray!20} \textbf{Model} & \textbf{Mean} & \textbf{LO} & \textbf{Math}  & \textbf{2D-G} & \textbf{3D-G} & \textbf{VI} & \textbf{TM} & \textbf{SR} & \textbf{CO}  \\ 
            \midrule
            \rowcolor{yellow!20} \multicolumn{10}{c}{\textbf{Open-Source LMMs}} \\
            LLaVA-1.6-7B~\cite{liu2024llavanext} & 19.45 & 24.22	& 20.34	& 17.92	& 15.83	& 20.00 &	18.23	& 17.82	& 18.42 \\ 
            Deepseek-vl-7b-chat~\cite{bi2024deepseek} & 22.17 & 19.53 & 20.30 & 22.25	& 27.39	& 35.56 & 23.72	& 24.75	& 15.79 \\
            Qwen2-VL-72B-Instruct~\cite{Qwen2VL} & 26.38 & 24.74 & 24.40 & 28.60 &	27.39 &24.44 &26.93	&32.67 &23.68 \\
            QVQ-72B-Preview~\cite{qvq-72b-preview} & 26.94 & 28.91	& 25.59	&29.23 &26.38 &26.67 &25.43	&22.77	&34.21 \\

            Qwen2.5-VL-7B-Instruct~\cite{bai2025qwen2} & 25.90 & 25.95 & 24.46 & 23.56 & 29.14 &25.53 &27.47	&27.96 &26.31 \\

            \rowcolor{cyan!20}
            Qwen2.5-VL-7B-Instruct (RL) & \textbf{30.77} & \textbf{30.28} & \textbf{29.48} & \textbf{32.05} & \textbf{30.15} & \textbf{27.66} & \textbf{32.05} & \textbf{34.74} & \textbf{39.47} \\

            \midrule
            \rowcolor{red!20} \multicolumn{10}{c}{\textbf{Proprietary LMMs}} \\
            GPT-4o~\cite{achiam2023gpt} & 26.87 & 25.52 & 25.70 & 28.32 & 27.64 &26.67 &25.69 &27.72	&50.00 \\
            Gemini-1.5-Pro-002~\cite{team2023gemini} & 26.86 & 19.53	&27.43	&28.03	&25.88	&24.44	&31.17	&25.74	&39.47 \\
            Claude-3.5-Sonnet~\cite{anthropic_claude3} & 27.49 & 23.41 
            &29.48 &26.60	&24.37	&35.56	&25.69	&27.72 &	42.11 \\
            
            Gemini-2.5-Pro~\cite{team2023gemini} &31.23	&33.33	&32.15	&30.96	&28.79	& 36.17	&27.23	&34.74	&42.10 \\
            
            Claude-3-7-Sonnet-Thinking~\cite{anthropic_claude3}	&31.55	& 32.57	&30.88	&33.69	&29.62	&31.91	&28.43	&35.59	&57.89 \\

           OpenAI-o3~\cite{jaech2024openai}	& 33.17	& 35.11	& 35.04	& 30.96	& 29.39	& 36.17	& 31.56	& 30.50	& 50.00 \\
           
            \midrule
            Human Performance & \textbf{51.27} & \textbf{61.36} & \textbf{45.03}  & \textbf{60.11} & \textbf{47.48} & \textbf{46.67} & \textbf{55.61}  & \textbf{36.63} & \textbf{65.79}  \\
            \bottomrule            
        \end{tabular}
    }
\end{table*}

\section{Experiments}

\subsection{Experimental Setup}
\label{sec:experimental setup}
\textbf{RL-trained Baseline} 
To set a baseline on the MM-IQ's training set, we choose the Qwen2.5-VL-7B-Instruct~\cite{bai2025qwen2} as the base model and utilize the GRPO algorithm that is implemented by the EasyR1 framework~\cite{zheng2025easyr1}. The entire training process is conducted on a cluster of 32 H20 GPUs. We set the global batch size to 128 and the rollout batch size to 512 to ensure stable gradient estimation across 4 distributed nodes. The sampling temperature is set to 1.0, and the number of responses to each question is 14 in order to balance exploration and exploitation during training. The learning rate is fixed at 1.0e-6 by default. The maximum response length during training is constrained to 4,096. The model is trained for a total of 25 episodes. The reward function used in the training process is a weighted sum of two components: format (with a weight of 0.1) and accuracy (with a weight of 0.9). For both format and accuracy, a correct response (via regular expression checking) is assigned a value of 1, while an incorrect response is assigned a value of 0. Additionally, a KL penalty with a coefficient of 1.0e-2 is applied to regularize the training process. Other unspecified parameters are set to the default values provided by the EasyR1 framework.

\textbf{Evaluation} We evaluated open-source and closed-source LMMs on the MM-IQ test set with zero-shot prompting and employed the same question prompt for all models. The few-shot prompting results will be included in the future version of MM-IQ since how to design appropriate multimodal prompts is still an open problem~\cite{yin2023survey,tai2024link}. For open-source LMMs, we selected widely used and state-of-the-art models, including QVQ-72B-Preview~\cite{qvq-72b-preview}, Qwen2-VL-72B-Instruct~\cite{Qwen2VL}, Deepseek-VL-7B-Chat~\cite{bi2024deepseek}, and LLaVA-1.6-7B~\cite{liu2024llavanext}. For closed-source LMMs, we adopted o3\cite{jaech2024openai}, Claude-3-7-Sonnet-Thinking\cite{anthropic_claude3}, Gemini-2.5-Pro\cite{team2023gemini}, GPT-4o-2024-08-06~\cite{achiam2023gpt}, Gemini-1.5-Pro-002~\cite{team2023gemini}, and Claude-3.5-Sonnet-2024-06-20~\cite{anthropic_claude3}. For a fair comparison, we employed the same settings and default hyperparameters for all LMMs (please refer to Table~\ref{tab:mllm_generating_params} in Appendix~\ref{appendix:experimental setup} for more details). Each model generates a single response to each problem in the dataset. Notably, o3, Claude-3-7-Sonnet-Thinking, Gemini-2.5-pro, and QVQ-72B-Preview are specifically designed for long CoT reasoning.
The evaluation process of LMMs consists of three steps: (1) response generation, (2) answer extraction, and (3) accuracy calculation. We extract the final answer using regular expression (regex) matching. For example, the final answer will be extracted from the response ``The correct answer is A.'' as ``A''. If there is no valid answer in the model’s response, it will be considered incorrect. \looseness-1

\subsection{Performance}

\textbf{Overall Performance:} 
Table~\ref{tab:performance} presents the performance of human participants and a diverse set of LMMs on the MM-IQ benchmark. Across all evaluated models, there remains a substantial gap between LMMs and human-level performance, with humans achieving a mean accuracy of 51.27\%, while the best-performing LMM, o3, attains 33.17\%. This performance gap is consistent across most paradigms, particularly in which requiring logical reasoning, 2D/3D geometry, and concrete object recognition, where human accuracy is markedly higher.

Focusing on open-source LMMs, we observe that models employing long chain-of-thought (CoT) strategies, such as QVQ-72B-Preview, achieve the highest mean accuracy (26.94\%) among their peers, with notable improvements in challenging categories like logical operation and mathematics. However, it is important to note that open-source models with short CoT, such as Qwen2-VL-72B-Instruct (26.38\%) and Qwen2.5-VL-7B-Instruct (25.90\%), deliver performance that is comparable to, and in some cases even surpasses, several proprietary models that also use short CoT (e.g., GPT-4o at 26.87\%, Gemini-1.5-Pro-002 at 26.86\%). This suggests that, in the absence of long CoT reasoning, the performance gap between open-source and proprietary LMMs is relatively small.

For all proprietary LMMs such as Gemini and o3, we observed a consistent phenomenon that models with long CoT outputs, such as Gemini-2.5-Pro (31.23\%), Claude-3-7-Sonnet-Thinking (31.55\%), and o3 (33.17\%), outperform their short counterparts. The performance gains are particularly evident in paradigms that require advanced reasoning capabilities, such as logical operations and mathematics. For instance, OpenAI-o3 achieves the highest accuracy in these paradigms. These results imply the effectiveness of long CoT in enhancing the abstraction and reasoning capabilities of LMMs, and indicate that more advances in this direction may help reduce the distance to human-level performance. \looseness-1

\begin{figure*}[h]
\centering
\includegraphics[width=0.99\textwidth]{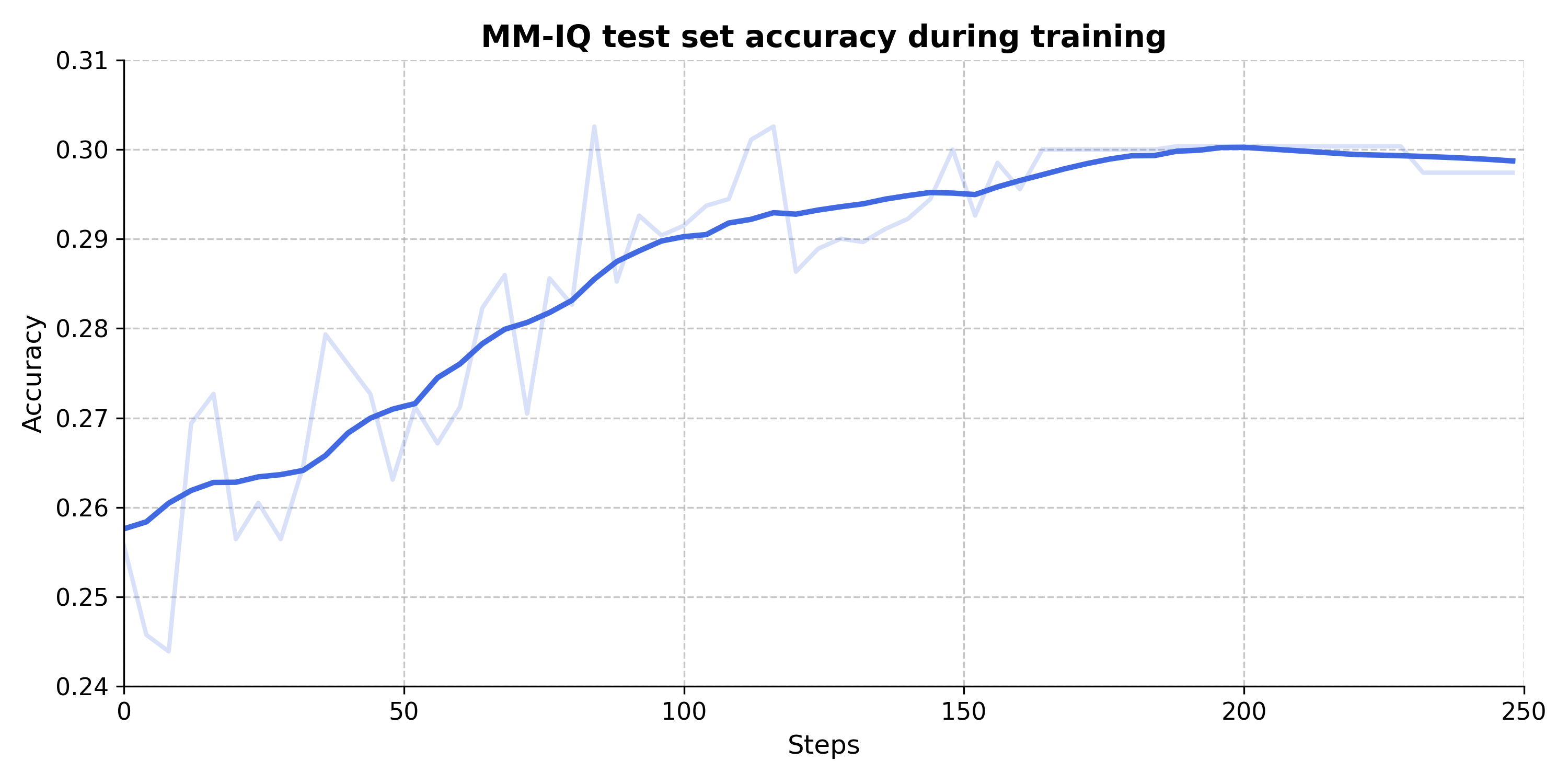    }
\caption{The test accuracy of Qwen2.5-VL-7B-Instruct on MM-IQ during RL training.\looseness-1}
\label{fig:Traing curve}
\end{figure*}

\textbf{Performance of RL-trained Baseline:} 
Figure~\ref{fig:Traing curve} depicts the accuracy curve on the MM-IQ test set of our RL-trained baseline with respect to training steps.
Notably, our baseline model demonstrates a significant improvement over its original version and other open-source LMMs, achieving performance comparable to Gemini-2.5-Pro, as shown in Table~\ref{tab:performance}. 
The improvement is consistent across all reasoning paradigms, with the RL-trained baseline achieving the highest scores among open-source models in logical operation (30.28\%), math (29.48\%), 2D-geometry (32.05\%), 3D-geometry (30.15\%), visual instruction (27.66\%), temporal movement (32.05\%), spatial relationship (34.74\%), and concrete object reasoning (39.47\%). These results suggest that RL can effectively incentivize the reasoning capabilities of LMMs, even for relatively small models. 

\section{Empirical Analysis}

\subsection{What Improvements does Reinforcement Learning Bring?}
\label{sec:What Improvements does Reinforcement Learning Bring?}

The performance analysis of the RL-trained baseline reveals notable improvements compared to its original version, prompting us to investigate which capabilities have been enhanced by RL. Specifically, the paradigms that require more complex reasoning gain higher improvements, such as 2D-Geometry (from 23.56\% to 32.05\%) and Spatial Relationship (from 27.96\% to 34.74\%). 

To understand the nature of these improvements, we examined specific instances where the RL-trained model outperformed its non-RL counterpart. One such example is illustrated in Figure~\ref{fig:baseline case} in Appendix~\ref{appenidx:Case Study of RL-trained Baseline}. This case demonstrates that the RL-trained model is better able to capture abstract rules within images and finer-grained relationships between objects, rather than merely focusing on superficial shape or positional information. This enhanced ability to abstract and reason allows the model to perform better overall, despite some remaining incorrect image descriptions and understandings. Across a wide range of such cases, we consistently observe that this enhanced pattern abstraction and reasoning capability is a key factor contributing to the overall performance gains, particularly in tasks that demand complex reasoning.

Additionally, the model has developed some reflective capabilities, although it may not always lead to correct analyses, as illustrated in Figure~\ref{fig:example of reflection} in Appendix~\ref{appenidx:Case Study of RL-trained Baseline}. This suggests that while the current model size may constrain its capabilities, the emergence of long CoT reasoning is a promising direction for future work. Therefore, we plan to train larger models using RL with MM-IQ training set to explore the frontier of reasoning in LMMs.


\subsection{Long CoT vs. Short CoT}

\begin{wraptable}{r}{0.55\textwidth}
\centering
\small 
\caption{Comparison of Reasoning Chain Length (Average Token Count)}
\label{tab:cot_length}
\begin{tabular}{lcc}
\toprule
\textbf{Model} & \textbf{Avg. Token Length} \\
\midrule
Gemini-1.5-Pro (Short CoT) & 98 \\
Gemini-2.5-Pro (Long CoT) & 2130 \\
\bottomrule
\end{tabular}
\end{wraptable}

As shown in Table~\ref{tab:performance}, models using long CoT reasoning consistently outperform those with short CoT reasoning, especially for proprietary models, which raises a question: in which aspects do long CoT models demonstrate enhanced capabilities compared to their short CoT counterparts? To systematically investigate this improvement, we compare Gemini-1.5-Pro (short CoT) and Gemini-2.5-Pro (long CoT) on the MM-IQ benchmark. Table~\ref{tab:cot_length} shows that the average token length of long CoT responses is over 20 times that of short CoT. Through a systematic analysis of a large number of model-generated examples, we observe that long CoT responses are not simply longer in length, but exhibit a fundamentally different reasoning process.

Specifically, long CoT models often adopt a combination of multiple thinking templates, including hypothesis generation, paradigm analysis, application, verification, etc. During this process, the model first proposes potential paradigms based on the input, then applies these hypotheses to the problem at hand, and finally verifies whether the resulting answer is consistent with all observed evidence. If inconsistencies are detected, the model revisits its previous steps, refines its hypotheses, and repeats the process until a satisfactory solution is found or the context length reaches a predefined threshold. \looseness-1

An example of this looped reasoning process is illustrated in Figure~\ref{fig:thinking process of long CoT} (Appendix~\ref{appendix:examples of top-performing models}), where the model not only identifies and follows human-like reasoning paradigms, but also actively checks the validity of the generated answer and makes necessary correction. In contrast, short CoT models typically provide a naive, single-pass response, lacking such a loop of self-correction and verification.

\subsection{Failure Analysis of Non-Reasoning LMMs on MM-IQ}

\begin{figure*}[h]
\centering
\includegraphics[width=0.95\textwidth]{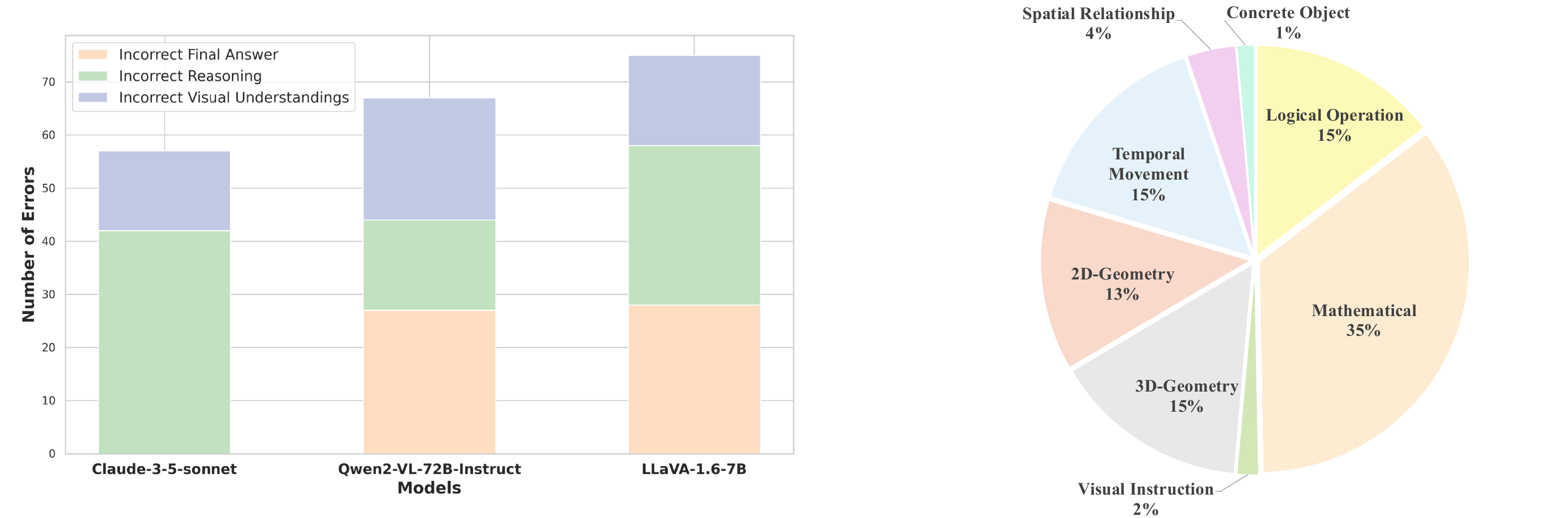}
\caption{\textbf{Left}: Distribution over different error types across three representative LMMs. \textbf{Right}: Quantitative distribution of reasoning paradigms in MM-IQ's test set. \looseness-1}
\label{fig:error_distribution_and_Data_distribution_pile}
\end{figure*}

Table~\ref{tab:performance} demonstrates that the highest accuracy of non-reasoning LMMs (Claude-3.5-Sonnet: 27.49\%) is almost equivalent to randomly guessing a correct answer among four options, which motivates us to ask: Does the top-performing non-reasoning LMM, e.g., Claude, actually possess the reasoning abilities required by AVR tasks? To investigate this, we selected three representative models: Claude-3.5-Sonnet, Qwen2-VL-72B-Instruct, and LLaVA-1.6-7B, and examined their generated wrong responses through human-in-the-loop evaluation. Specifically, we sampled a total of 90 predictions from each model for analysis. These 90 questions included 10 instances from each reasoning paradigm, as well as 20 from the Mathematical paradigm, given its larger proportion in the MM-IQ dataset (35.1\%).

\textbf{Response Style and Structure.}
We first examined the average length and style of the generated responses. Compared to LLaVA-1.6-7B and Qwen2-VL-72B-Instruct, the best LMM Claude-3.5-Sonnet produces longer and structured answers. Moreover, Claude-3.5-Sonnet’s responses begin with a detailed description of the given image and strategic problem-solving plans, followed by a discussion of each option to determine the correct answer. A visual example of Claude-3.5-Sonnet's response is illustrated in Figure~\ref{fig:example of Claude-3.5-sonnet's prediction} in Appendix~\ref{appendix:examples of top-performing models}. In contrast, LLaVA-1.6-7B and Qwen2-VL-72B-Instruct fail to generate such responses. These observations suggest that structured reasoning may enhance performance on MM-IQ.

\textbf{Error Typology.}
We examine each wrong response and categorize them into three types: incorrect paradigm reasoning, incorrect final answers, and incorrect visual understanding, examples of which can be found in Figure~\ref{fig:incorrect pattern reasoning}, Figure~\ref{fig:incorrect final answer}, and Figure~\ref{fig:incorrect visual understanding} in Appendix~\ref{appendix:exmaples of three error typologies}. As shown in Figure~\ref{fig:error_distribution_and_Data_distribution_pile}, incorrect paradigm reasoning forms a major part of failures (32.3\% on average). In these responses, we observe that LMMs often solve problems by using wrong rules or focusing on more superficial changes. These wrong rules include objects in the image becoming progressively more compact and complex. A corresponding visualized example is provided in Figure~\ref{fig:incorrect pattern reasoning}, where the red texts denote the incorrect reasoning due to the usage of infeasible rules.

We also observed that Qwen2-VL-72B-Instruct and LLaVA-1.6-7B frequently make errors by directly providing the final answer without any thinking.
To further investigate whether the absence of thinking processes is a critical factor, we remove all responses' thinking processes and reevaluate their accuracy. For top-performing models, such as Qwen2-VL-72B-Instruct, only generating the final answer results in a performance drop of 4.7\% (from 26.9\% to 22.5\%) on average. Conversely, for LLaVA-1.6-7B, it leads to an improvement of 2.8\% (from 19.4\% to 22.2\%), implying that larger and stronger LMMs benefit more from long CoT reasoning.

A deeper analysis of visual understanding errors reveals that all three models struggle with paradigms involving complex visual content, such as logical operations, temporal movement, and spatial relationships (see Figure~\ref{fig:incorrect visual understanding across eight reasoning paradigms} in Appendix~\ref{appendix:analysis of incorrect visual understanding}). Moreover, there is an inverse correlation between a model’s overall performance and the proportion of its visual understanding errors. For instance, Claude-3.5-Sonnet performs poorly on temporal movement and spatial relationship reasoning paradigms, and also performs worse on visual understanding of both paradigms. This underscores the necessity of enhancing the models' perceptual capacity to accurately interpret complex visual paradigms, thereby improving LMMs' reasoning capabilities. Due to limitated space, more discussions can be found in Appendix~\ref{appendix:analysis of incorrect visual understanding}.

In summary, our failure analysis of LMMs on the MM-IQ dataset highlights several critical points for further research and improvement in multimodal abstract reasoning: 1) \textbf{Structured Response Generation}: Models like Claude-3.5-Sonnet, which produce longer and more structured responses, tend to perform better, suggesting that enhancing the ability to generate structured and detailed reasoning chains can improve accuracy. 2) \textbf{Abstract Pattern Recognition}: A significant portion of errors stems from incorrect reasoning due to reliance on simpler rules. Improving models' ability to identify and apply high-level abstract paradigms is essential. 3) \textbf{Explanatory Vs. Concise Answers}: The presence of detailed explanations can improve performance in stronger models but may not benefit weaker ones, highlighting the nuanced role of explanatory reasoning in model accuracy. 4) \textbf{Visual Understanding}: All models exhibit poor performance on complex visual paradigms, such as logical operations and spatial relationships, indicating a need for enhanced perceptual capabilities to accurately interpret intricate visual details. Addressing these challenges is crucial for advancing the reasoning capabilities of LMMs.

\vspace{-0.5em}
\section{Conclusion}
\vspace{-0.5em}
\label{sec:conc}
We propose MM-IQ, a comprehensive benchmark for evaluating the abstract visual reasoning of LMMs, which comprises a large-scale training set with 4,776 visual reasoning problems and 2,710 meticulously curated test items across 8 distinct reasoning paradigms, enabling a rigorous assessment of LMMs' abstraction and reasoning capabilities. Experimental results reveal striking limitations in current state-of-the-art LMMs, with the leading models achieving only slightly above the accuracy of random guessing, far behind human performance. We conduct a thorough failure analysis that identifies several key points for improvement, including structured reasoning, abstract pattern recognition, visual understanding, and inference-time scaling. MM-IQ is expected to complement existing multimodal benchmarks and provide a valuable resource for steering progress in multimodal research and promoting the advancements of AGI.

\textbf{Limitation and broader impacts} Due to limited resources, we cannot train larger models using RL on MM-IQ as strong baselines, and plan to extend this in future work. Regarding broader impacts of MM-IQ, it is a fundamental research project for AI and LMMs, and we anticipate no adverse societal consequences beyond those generally associated with other widely used datasets and benchmarks.

\clearpage


\bibliography{custom}

\bibliographystyle{plainnat}
\clearpage

\appendix


\section{Appendix}

\subsection{Experimental Setup}
\label{appendix:experimental setup}
\begin{table*}[h]
    \centering
    \caption{Generating parameters for various LMMs.}
    \label{tab:mllm_generating_params}
    \resizebox{\textwidth}{!}{
        \begin{tabular}{l|>{\raggedright\arraybackslash}p{0.7\textwidth}}
            \toprule
            \textbf{Model} & \textbf{Generation Setup} \\
            \midrule
            Claude-3.5-Sonnet-2024-06-20 & \texttt{temperature = 1.0, output\_token\_limit = 8,192, top\_p = 1.0} \\
            \midrule
            GPT-4o-2024-08-06 & \texttt{temperature = 1.0, output\_token\_limit = 16,384, top\_p = 1.0} \\
            \midrule
            Gemini-1.5-Pro-002 & \texttt{temperature = 1.0, output\_token\_limit = 8,192} \\
            \midrule
            DeepSeek-vl-7b-chat & \texttt{temperature = 1.0, output\_token\_limit = 2,048, do\_sample = False, top\_p = 1.0} \\
            \midrule
            LLaVA-1.6-7B & \texttt{temperature = 0, output\_token\_limit = 2,048} \\
            \midrule
            Qwen2-VL-72B-Instruct & \texttt{temperature = 1.0, output\_token\_limit = 8,192, top\_p = 0.001, top\_k = 1, do\_sample = True,} \\
            & \texttt{repetition\_penalty = 1.05} \\
            
            \midrule
            QVQ-72B-Preview & \texttt{temperature = 0.01, output\_token\_limit = 8,192, top\_p = 0.001, top\_k = 1, do\_sample = True,} \\
            & \texttt{repetition\_penalty = 1.0} \\

           \midrule
            Claude-3-7-Sonnet-Thinking-20250219 & \texttt{temperature = 1.0, output\_token\_limit = 20,000, top\_p = 1.0} \\

           \midrule
            Gemini-2.5-pro-preview-03-25 & \texttt{temperature = 1.0, output\_token\_limit = 20,000, top\_p = 0.95, top\_k = 64} \\
            
           \midrule
            o3-2025-03-01-preview & \texttt{temperature = 1.0, output\_token\_limit = 20,000, top\_p = 1.0} \\
            \bottomrule
        \end{tabular}
    }
\end{table*}

\subsection{Analysis of Incorrect Visual Understanding}
\label{appendix:analysis of incorrect visual understanding}

\begin{figure*}[h]
\centering
\includegraphics[width=1\textwidth]{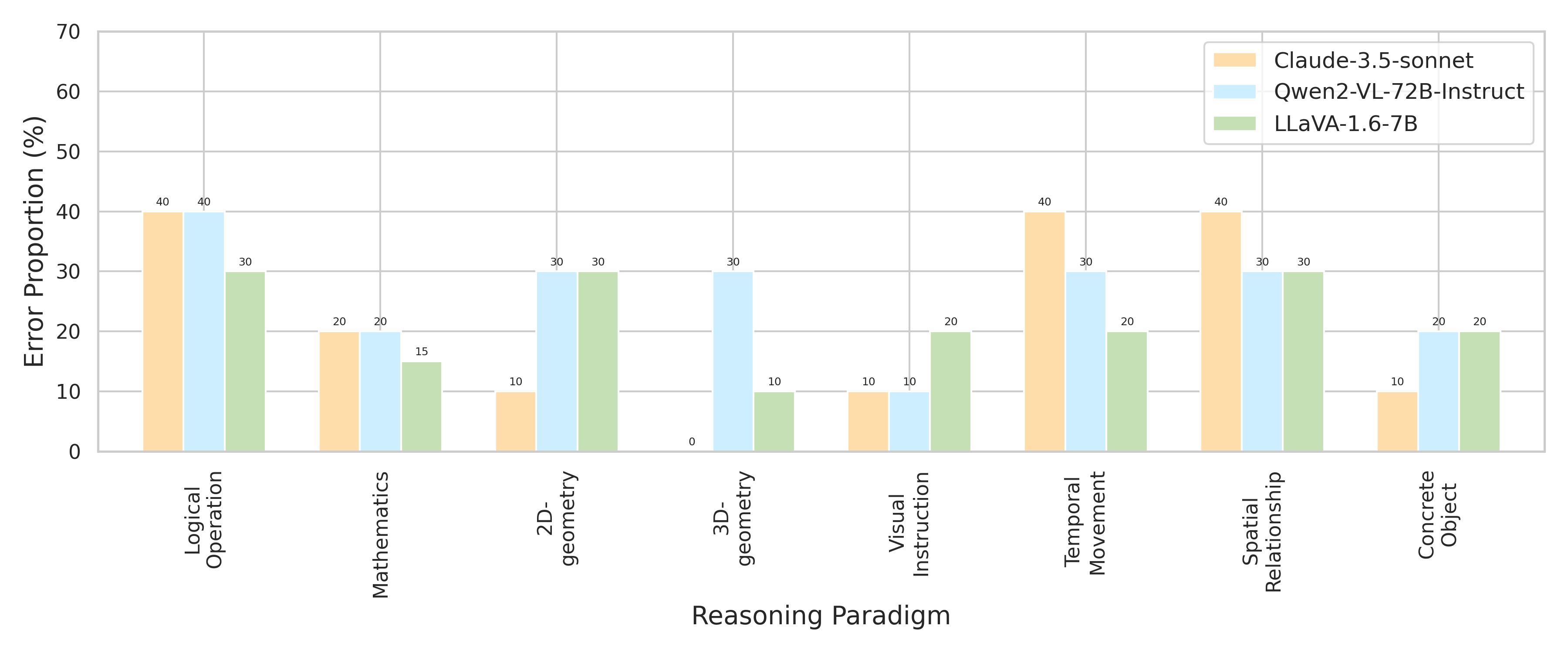}
\caption{\textbf{Proportions of incorrect visual understanding across eight reasoning paradigms.}\looseness-1}
\label{fig:incorrect visual understanding across eight reasoning paradigms}
\end{figure*}

A more granular analysis of visual understanding errors across different reasoning paradigms (see Figure~\ref{fig:incorrect visual understanding across eight reasoning paradigms}) reveals several key patterns. For tasks involving logical operations, temporal movement, and spatial relationships, all three models exhibit a high proportion of visual understanding errors (up to 40\% for Claude-3.5-sonnet in these categories). This suggests that, for these complex paradigms, the primary bottleneck lies in the models' ability to accurately perceive and interpret visual information, rather than in subsequent reasoning steps.

In contrast, for mathematical and visual instruction paradigms, the proportion of visual understanding errors is much lower (typically below 20\%), indicating that errors in these categories are more likely due to reasoning or calculation failures rather than perceptual limitations. For concrete object recognition, all models show relatively low visual understanding error rates, reflecting their strong performance in basic object identification.

Comparing the three models, Claude-3.5-sonnet tends to have a higher proportion of visual understanding errors in the most challenging paradigms (such as logical operation, temporal movement, and spatial relationship), while Qwen2-VL-72B-Instruct and LLaVA-1.6-7B show a more balanced distribution of error types across paradigms. This highlights that, although some models may achieve higher overall accuracy, their performance in complex visual reasoning tasks is still constrained by perceptual limitations.

In summary, the analysis of error distributions and types reveals that visual understanding remains a major challenge for LMMs, especially in tasks involving abstract, dynamic, or relational visual reasoning. Addressing these weaknesses is crucial for advancing the overall reasoning capabilities of multimodal models.

\subsection{Examples of Three Error Typologies.}
\label{appendix:exmaples of three error typologies}

\begin{figure*}[h!]
    \centering
    \begin{tcolorbox}[colframe=RoyalBlue, colback=white]
    \textbf{Prompt:} Which of the following options conforms to the given pattern of changes in the figures? \\
    \\
    \begin{minipage}{0.8\textwidth}
        \centering
    \includegraphics[scale=0.6]{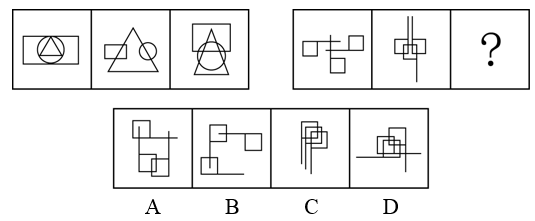}
    \end{minipage}
    \\
    \\
    \textbf{LMM's Response:} 
    \begin{tcolorbox}[colframe=RoyalBlue, colback=gray!10]
To solve this problem, we need to analyze the pattern in the top row and then find which option in the bottom row follows that pattern.

In the top row, we see:
1. A circle inside a triangle inside a square
2. The same shapes, but the triangle is cut off on one side
3. The same shapes, but now the circle is cut off
4. A new pattern of intersecting squares
5. The same intersecting squares, but with an additional square added

\textcolor{red}{The pattern seems to be adding complexity or elements as we move from left to right.}

Now, looking at the options A, B, C, and D:

A: Shows two intersecting squares

B: Shows three intersecting squares

C: Shows four intersecting squares

D: Shows five intersecting squares

\textcolor{red}{Following the pattern of increasing complexity and number of elements, the correct answer should be D.} It shows the most complex arrangement with five intersecting squares, which follows the progression we see in the top row.

Therefore, the answer that best conforms to the given pattern is option D.
\end{tcolorbox}

\textbf{Final Answer:} \textbf{D} \textcolor{red}{{\Large \texttimes}}  \\
\textbf{Ground Truth:} \textbf{C} 
    \end{tcolorbox}
    \caption{An example of incorrect paradigm reasoning. The \textcolor{red}{red parts} are the incorrect reasoning due to wrongly recognizing simpler rules. The correct reasoning solution is that the intersection in the third panel is the sum of the intersections in the first two panels in the same box.}
    \label{fig:incorrect pattern reasoning}
\end{figure*}

\begin{figure*}[h!]
    \centering
    \begin{tcolorbox}[colframe=RoyalBlue, colback=white]
    \textbf{Prompt:} Choose the only option from the four figures to fill in the question mark, so that it presents a certain regularity. \\
    \\
    \begin{minipage}{0.8\textwidth}
        \centering
    \includegraphics[scale=0.2]{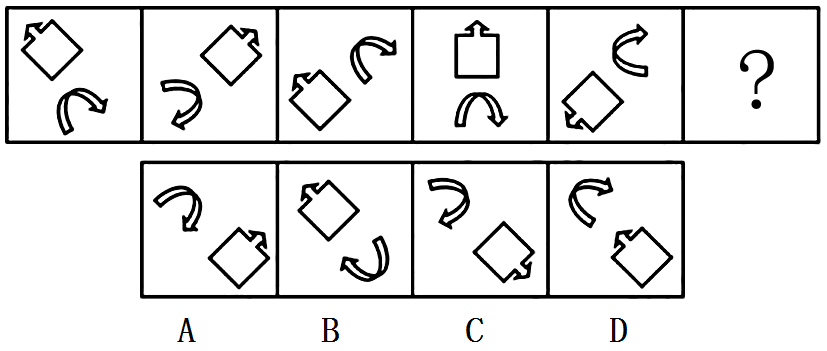}
    \end{minipage}
    \\
    \\
    \textbf{LMM's Response:} 
    \begin{tcolorbox}[colframe=RoyalBlue, colback=gray!10]
The only option that presents a certain regularity is B.
\end{tcolorbox}

\textbf{Final Answer:} \textbf{B} \textcolor{red}{{\Large \texttimes}}  \\
\textbf{Ground Truth:} \textbf{A} 
    \end{tcolorbox}
    \caption{An example of incorrect final answers.}
    \label{fig:incorrect final answer}
\end{figure*}

\begin{figure*}[h!]
    \centering
    \begin{tcolorbox}[colframe=RoyalBlue, colback=white]
    \textbf{Prompt:} Choose the most appropriate one from the given four options to present a certain regularity: \\
    \\
    \begin{minipage}{0.8\textwidth}
        \centering
    \includegraphics[scale=0.55]{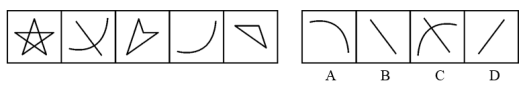}
    \end{minipage}
    \\
    \\
    \textbf{LMM's Response:} 
    \begin{tcolorbox}[colframe=RoyalBlue, colback=gray!10]
To solve this problem, we need to analyze the pattern in the given sequence of shapes and then determine which of the options A, B, C, or D best continues that pattern.

Let's examine the sequence:

1. Star (5-pointed)

2. Shape with two curved lines

3. Shape with two straight lines forming an acute angle

4. Shape with one curved line (quarter circle)

5. \textcolor{red}{Shape with one straight line}

The pattern seems to be a progression from more complex shapes to simpler ones, reducing the number of lines or points:  - From 5 points (star) to 2 lines, then to 1 line  - From curved lines to straight lines
Following this pattern, the next shape should be even simpler, possibly just a single straight line or a single point.

Looking at the options:

A: Curved line

B: Straight line

C: \textcolor{red}{Two intersecting straight lines}

D: \textcolor{red}{Single point}

Option B (straight line) or D (single point) would be the most logical continuations of the pattern. However, since we've already seen a straight line in the sequence, the single point (Option D) represents a further simplification and reduction, making it the most appropriate choice to continue the observed pattern.

Therefore, the best answer is D.
\end{tcolorbox}

\textbf{Final Answer:} \textbf{D} \textcolor{red}{{\Large \texttimes}}  \\
\textbf{Ground Truth:} \textbf{B} 
    \end{tcolorbox}
    \caption{An example of incorrect visual understanding. The \textcolor{red}{red parts} are the incorrect visual understanding. }
    \label{fig:incorrect visual understanding}
\end{figure*}

This section presents visualized examples of three error typologies-including incorrect
paradigm reasoning, incorrect final answers, and incorrect visual understanding, which
can be found in Figure~\ref{fig:incorrect pattern reasoning}, Figure~\ref{fig:incorrect final answer}, and Figure~\ref{fig:incorrect visual understanding}. 

\clearpage

\subsection{Examples of Reasoning Paradigms of MM-IQ}
\label{appendix:example of reasoning paradigms}

\begin{figure*}[h!]
    \centering
    \begin{tcolorbox}[colframe=RoyalBlue, colback=white]
    \textbf{Prompt:} The option that best fits the given pattern of figures is ( ).\\
    \\
    \begin{minipage}{0.6\textwidth}
        \centering
    \includegraphics[scale=0.19]{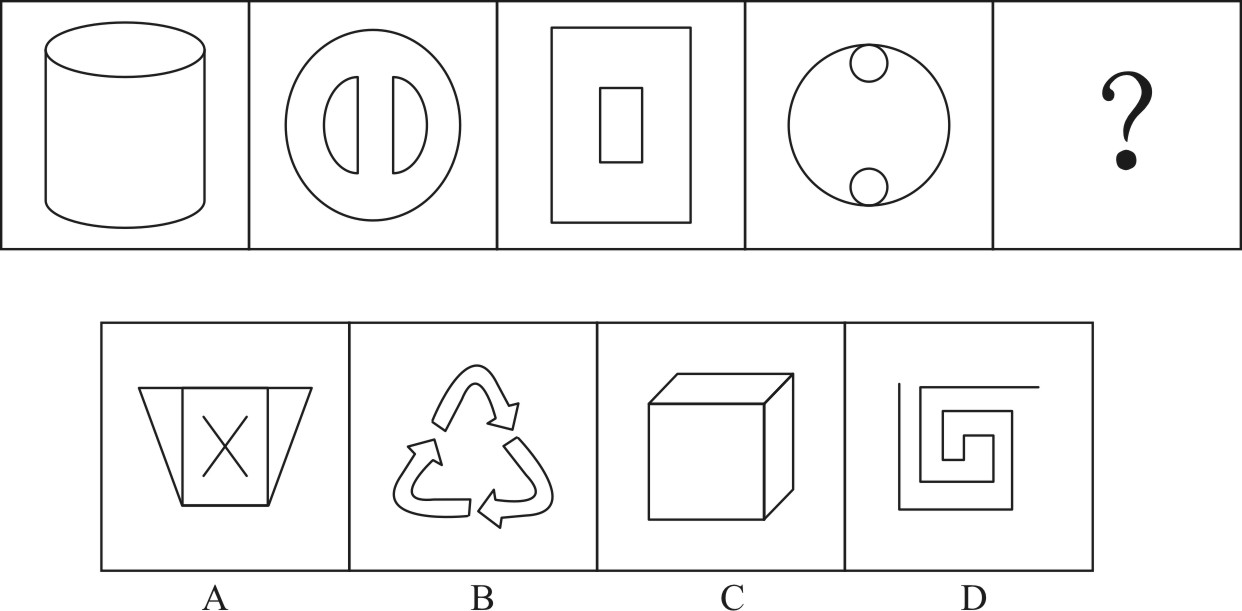}
    \end{minipage}
    \\
    \\
    \textbf{LMM's Response:} 
    \begin{tcolorbox}[colframe=RoyalBlue, colback=gray!10]
To determine the option that best fits the given pattern of figures, let's analyze the symmetry of each figure in the sequence:

The first panel is a cylinder, which has rotational symmetry around its vertical axis.
The second panel is a circle with two vertical lines inside, which has vertical symmetry.
The third panel is a rectangle inside a square, which have both vertical and horizontal symmetry.
The fourth panel is a circle with two small circles inside, which has rotational symmetry around its center.
Now, let's consider the options:

A. A trapezoid with an X inside, which has vertical symmetry.

B. A recycling symbol, which has no symmetry.

C. A cube, which has multiple planes of symmetry.

D. A spiral, which has no symmetry.

Based on the symmetry analysis, the option that best fits the pattern of figures is A, as it has rotational symmetry similar to the other figures in the sequence.
\end{tcolorbox}

\textbf{Ground Truth:} \textbf{A}
    \end{tcolorbox}
    \caption{A visualized example of 2D-geometry paradigm.}
    \label{fig:2D geometry reasoning}
\end{figure*}

\begin{figure*}[h!]
    \centering
    \begin{tcolorbox}[colframe=RoyalBlue, colback=white]
    \textbf{Prompt:} Choose the most appropriate option from the given four options to present a certain regularity:\\
    \\
    \begin{minipage}{0.7\textwidth}
        \centering
    \includegraphics[scale=0.38]{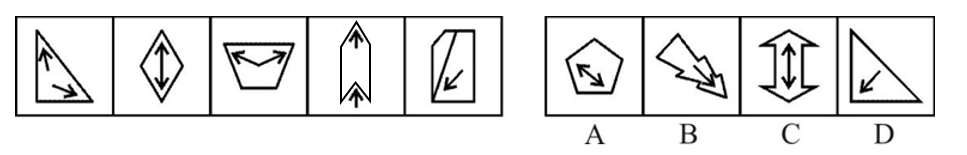}
    \end{minipage}
    \\
    \\
    \textbf{LMM's Response:} 
    \begin{tcolorbox}[colframe=RoyalBlue, colback=gray!10]
By looking at the images, we notice that each picture has arrows, so we focus on their direction. When we examine the figure in the question from left to right, we see that all the arrows point to acute angles. Among the four options, only option B meets the requirement that the arrows point to acute angles.

Therefore, option B is the correct choice.
\end{tcolorbox}

\textbf{Ground Truth:} \textbf{B}
    \end{tcolorbox}
    \caption{A visualized example of visual instruction paradigm.}
    \label{fig:visual instruction reasoning}
\end{figure*}

\begin{figure*}[h!]
    \centering
    \begin{tcolorbox}[colframe=RoyalBlue, colback=white]
    \textbf{Prompt:} Choose the most appropriate option from the given four options to present a certain regularity:\\
    \\
    \begin{minipage}{0.7\textwidth}
        \centering
    \includegraphics[scale=0.6]{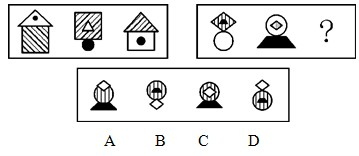}
    \end{minipage}
    \\
    \\
    \textbf{LMM's Response:} 
    \begin{tcolorbox}[colframe=RoyalBlue, colback=gray!10]
As can be seen from the figure, the number of elements is the same, but the styles are different, mainly indicating that the positions and methods of placing multiple elements vary. We can identify the pattern in the three figures from the left box and then apply this pattern to the right box. All the figures in the square box on the left are composed of solid circles, triangles, and quadrilaterals. The positions of the solid circles are upper inner, lower, and lower inner, respectively; the positions of the triangles are upper outer, upper inner, and upper, respectively; and the positions of the quadrilaterals are lower, upper outer, and lower outer, respectively. By applying this pattern to the square box on the right, the figure at the question mark should be a solid trapezoid in the lower inner position, a rhombus in the upper position, and a circle in the lower outer position, which corresponds to option D.

Therefore, the correct option is option D.
\end{tcolorbox}

\textbf{Ground Truth:} \textbf{D}
    \end{tcolorbox}
    \caption{A visualized example of spatial relationship paradigm.}
    \label{fig:spatial relationship reasoning}
\end{figure*}

\begin{figure*}[h!]
    \centering
    \begin{tcolorbox}[colframe=RoyalBlue, colback=white]
    \textbf{Prompt:} Choose the most appropriate option from the given choices: \\
    A: 4;  B: 5;  C: 6;  D: 7.
    \\
    \begin{minipage}{0.5\textwidth}
        \centering
    \includegraphics[scale=0.8]{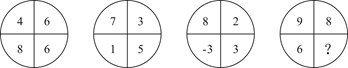}
    \end{minipage}
    \\
    \\
    \textbf{LMM's Response:} 
    \begin{tcolorbox}[colframe=RoyalBlue, colback=gray!10]
By observing the numbers in the figure, we notice the following patterns: 6 - 4 = 8 - 6, 5 - 7 = 1 - 3, and 3 - 8 = -3 - 2. The pattern is that the number in the bottom right corner minus the number in the top left corner equals the number in the bottom left corner minus the number in the top right corner.

Using this pattern, we can set up the equation: ? - 9 = 6 - 8. Solving for ?, we find that ? = 7. 

Therefore, we choose option D.
\end{tcolorbox}

\textbf{Ground Truth:} \textbf{D}
    \end{tcolorbox}
    \caption{A visualized example of mathematics paradigm.}
    \label{fig:arithmetic visual reasoning}
\end{figure*}

\begin{figure*}[h!]
    \centering
    \begin{tcolorbox}[colframe=RoyalBlue, colback=white]
    \begin{minipage}[t]{0.37\textwidth}
        \textbf{Prompt:} Choose the most appropriate option from the given four choices to fill in the question mark, so that it presents a certain regularity:\\
        \\
        \hfill
        \begin{minipage}{0.38\textwidth}
            \centering
            \includegraphics[scale=0.75]{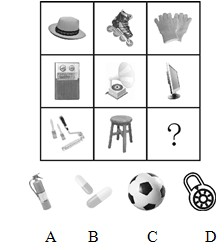}
        \end{minipage}
    \end{minipage}
    \hfill
    \begin{minipage}[t]{0.6\textwidth}
        \textbf{LMM's Response:} 
        \begin{tcolorbox}[colframe=RoyalBlue, colback=gray!10]
To solve this puzzle, we need to identify the pattern or regularity in the given grid.

Looking at the grid:

1. The first row contains a hat, roller skates, and gloves, which are all wearable items.

2. The second row contains a radio, a gramophone, and a television, which are all related to audio-visual media.

3. The third row contains tools (screwdrivers), a stool, and the missing item.

The pattern seems to be categorization based on function or use. The third row appears to be related to household or utility items.

Given the options:
- A: Fire extinguisher
- B: Pills
- C: Soccer ball
- D: Padlock

The most appropriate choice to complete the pattern of household or utility items is the padlock (D).
        \end{tcolorbox}
        \textbf{Ground Truth:} \textbf{D}
    \end{minipage}
    \end{tcolorbox}
    \caption{A visualized example of concrete object paradigm.}
    \label{fig:concrete object reasoning}
\end{figure*}

\begin{figure*}[h!]
    \centering
    \begin{tcolorbox}[colframe=RoyalBlue, colback=white]
    \textbf{Prompt:} The one that matches the top view is:\\
    \\
    \begin{minipage}{0.5\textwidth}
        \centering
    \includegraphics[scale=0.13]{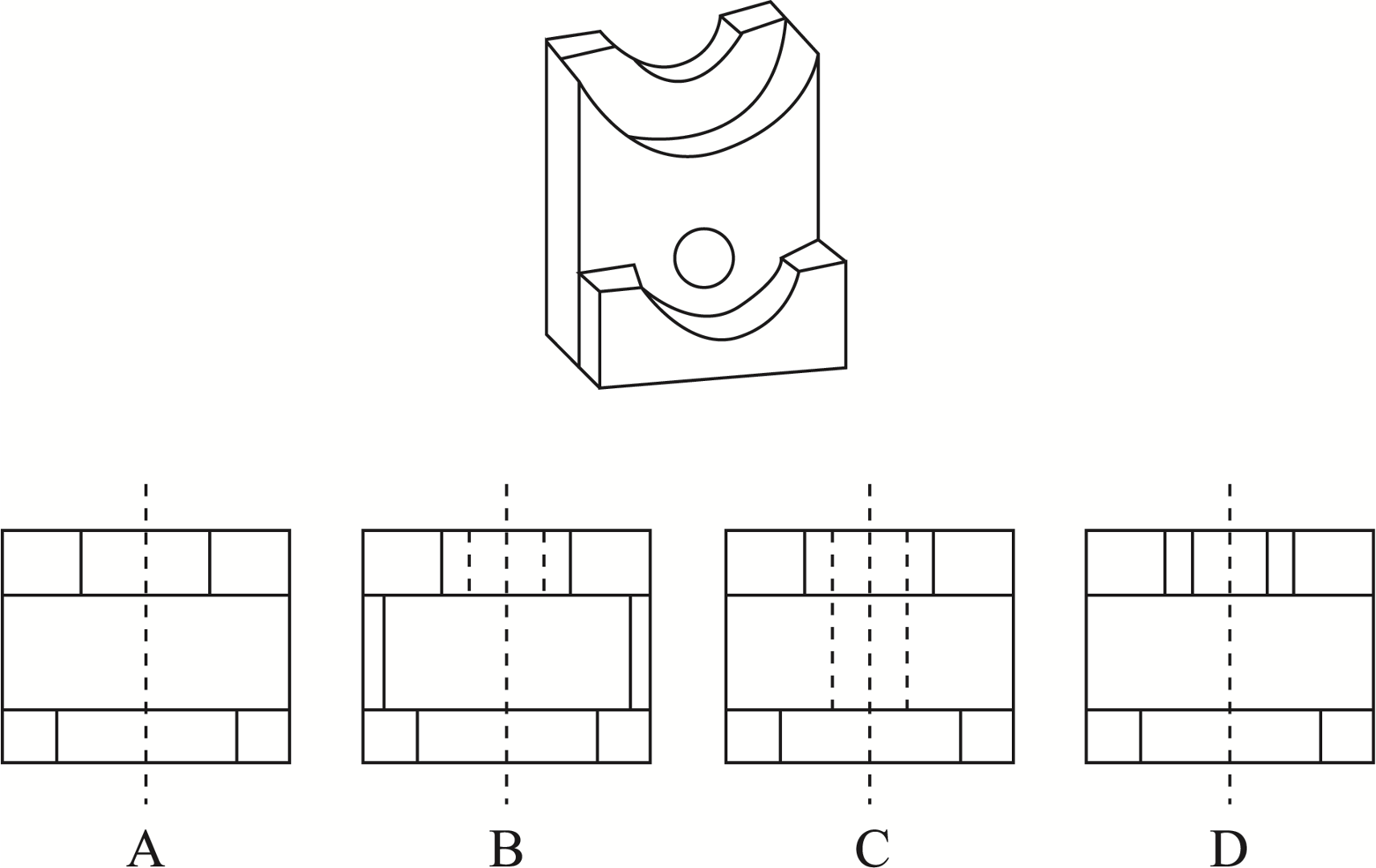}
    \end{minipage}
    \\
    \\
    \textbf{LMM's Response:} 
    \begin{tcolorbox}[colframe=RoyalBlue, colback=gray!10]
The top view is the view obtained by orthographic projection from the top of the object, also called the top view. Analysis options:

A: There is a hollow cylinder in the three-dimensional figure that is not reflected, so it is excluded;

B: The hollow cylinder in the three-dimensional figure is only partially reflected, which is incomplete, so it is excluded;

C: There is a hollow cylinder in the middle of the bottom of the three-dimensional figure. The visible lines are represented by solid lines, and the invisible but real lines are represented by dotted lines. This is correct;

D: When observing the top row of the three-dimensional figure from top to bottom, there should be only 3 solid line surfaces, so it is excluded.

Therefore, the final choice is option C.
\end{tcolorbox}

\textbf{Ground Truth:} \textbf{C}
    \end{tcolorbox}
    \caption{A visualized example of 3D-geometry reasoning.}
    \label{fig:3D geometry reasoning}
\end{figure*}

\begin{figure*}[h!]
    \centering
    \begin{tcolorbox}[colframe=RoyalBlue, colback=white]
    \begin{minipage}[t]{0.45\textwidth}
        \textbf{Prompt:} Choose the most appropriate option from the given four choices to fill in the question mark, so that it presents a certain regularity:\\
        \\
        \hfill
        \begin{minipage}{0.8\textwidth}
            \centering
            \includegraphics[scale=.22]{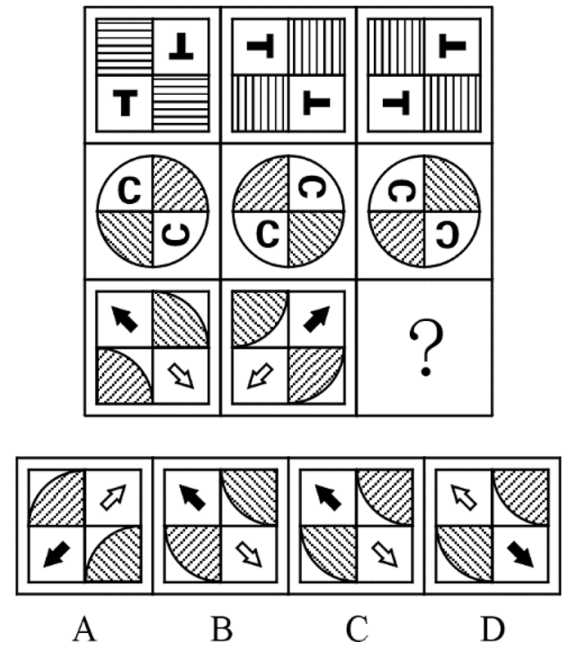}
        \end{minipage}
    \end{minipage}
    \hfill
    \begin{minipage}[t]{0.45\textwidth}
        \textbf{LMM's Response:} 
        \begin{tcolorbox}[colframe=RoyalBlue, colback=gray!10]
By observing the images, we can see that in the first row, image 1 is rotated 90 degrees clockwise to produce image 2, and then image 2 is flipped to the right to create image 3. The second row follows the same pattern.

Applying this same logic to the third row, we flip image 2 to the right, which allows us to obtain option C.

Therefore, the final choice is option C.

        \end{tcolorbox}
        \textbf{Ground Truth:} \textbf{C}
    \end{minipage}
    \end{tcolorbox}
    \caption{A visualized example of temporal movement paradigm.}
    \label{fig:temporal movement reasoning}
\end{figure*}

This section presents visualized examples of various reasoning paradigms—including visual instruction, spatial relationship, mathematics, concrete object, 2D-geometry, 3D-geometry reasoning, and temporal movement. Each example (see Figures~\ref{fig:visual instruction reasoning},~\ref{fig:spatial relationship reasoning},~\ref{fig:arithmetic visual reasoning},~\ref{fig:concrete object reasoning},~\ref{fig:2D geometry reasoning}, ~\ref{fig:3D geometry reasoning}, and~\ref{fig:temporal movement reasoning}) illustrates the corresponding paradigm to facilitate better understanding of the MM-IQ benchmark.

\begin{figure*}[h!]
    \centering
    \begin{tcolorbox}[colframe=RoyalBlue, colback=white]
    \begin{minipage}[t]{0.45\textwidth}
        \textbf{Prompt:} Choose the most appropriate option from the given four choices to fill in the question mark, so that it presents a certain regularity:\\
        \\
        \hfill
        \begin{minipage}{0.8\textwidth}
            \centering
            \includegraphics[scale=.25]{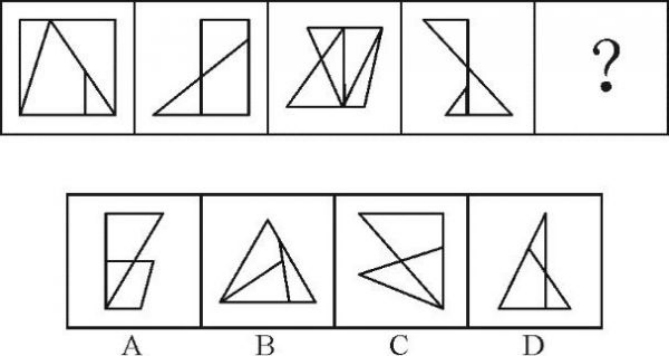}
        \end{minipage}
    \end{minipage}
    \hfill
    \begin{minipage}[t]{0.46\textwidth}
        \textbf{LMM's Response:} 
        \begin{tcolorbox}[colframe=RoyalBlue, colback=gray!10]
        Based on the pattern observed, each figure in the sequence has 7 intersections. After analyzing the options, only option A meets this requirement.

        Therefore, the correct answer is option A.
        \end{tcolorbox}
        \textbf{Ground Truth:} \textbf{A}
    \end{minipage}
    \end{tcolorbox}
    \caption{A visualized example of mathematics paradigm.}
    \label{fig:quantity reasoning}
\end{figure*}

\subsection{Examples of Top-performing Models}
\label{appendix:examples of top-performing models}
\begin{figure*}[h!]
    \centering
    \begin{tcolorbox}[colframe=RoyalBlue, colback=white]
    \textbf{Prompt:} The option that best fits the given pattern of figures is ( ).\\
    \\
    \begin{minipage}{0.5\textwidth}
        \centering
    \includegraphics[scale=0.23]{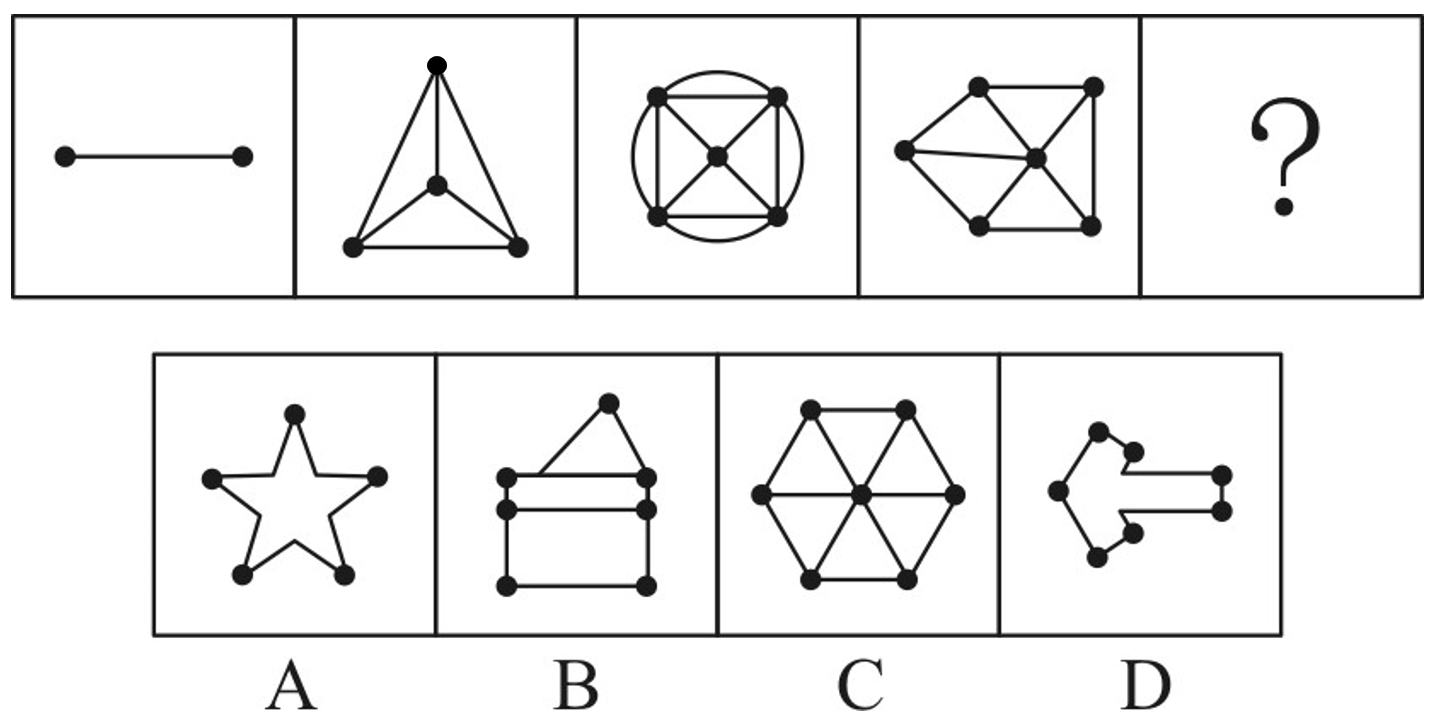}
    \end{minipage}
    \\
    \\
    \textbf{Response:} 
    \begin{tcolorbox}[colframe=RoyalBlue, colback=gray!10]
To solve this problem, we need to analyze the pattern in the given sequence of figures and then determine which of the options (A, B, C, or D) best continues that pattern.

Looking at the sequence:

1. We start with a simple line connecting two points.

2. Then we have a triangle with 3 points.

3. Next is a shape with 4 points, forming a square with crossed lines inside.

4. The fourth shape has 5 points, forming an irregular pentagon.

The pattern seems to be increasing the number of points by one each time, while also increasing the complexity of the connections between these points.

Following this pattern, the next shape should have 6 points.

Looking at the options:

A. A star shape with 5 points - doesn't fit the pattern as it doesn't increase to 6 points.

B. A house-like shape with 6 points - fits the number of points, but the connections are simpler than the previous shapes.

C. A hexagon with interconnected points - this fits both the number of points (6) and the complexity of connections.

D. An irregular shape with 6 points - fits the number of points, but the connections don't follow the increasing complexity.

Therefore, the option that best fits the given pattern is C. It has 6 points, continuing the increasing point count, and its interconnected structure follows the pattern of increasing complexity in how the points are connected.
\end{tcolorbox}

\textbf{Final Answer:} \textbf{C}  \textcolor{green}{{\Large \checkmark}}
    \end{tcolorbox}
    \caption{An example of Claude-3.5-Sonnet's prediction.}
    \label{fig:example of Claude-3.5-sonnet's prediction}
\end{figure*}

\begin{figure*}[h!]
    \centering
    \begin{tcolorbox}[colframe=RoyalBlue, colback=white]
    \textbf{Prompt:} Choose the most appropriate one from the given four options to present a certain regularity:\\
    \\
    \begin{minipage}{0.9\textwidth}
        \centering
    \includegraphics[scale=0.25]{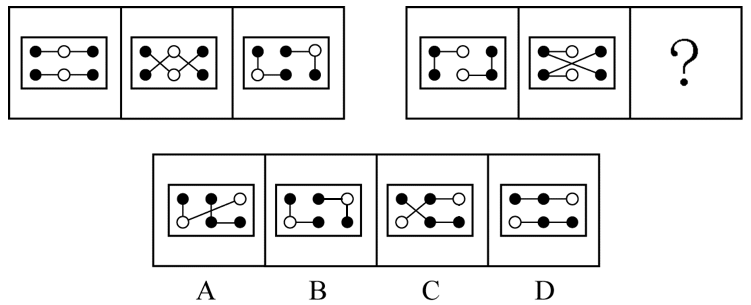}
    \end{minipage}
    \\
    \\
    \textbf{Gemini-1.5-Pro's Response (Short CoT):} 
    \begin{tcolorbox}[colframe=RoyalBlue, colback=gray!10]
Here's how to solve this matrix reasoning puzzle:

1. **Connections Change:**  Notice how the lines connecting the circles change with each step. They seem to rotate or shift.

2. **Color Change:** The black and white circles also change positions.  There's a sort of flipping or swapping happening.

3. **Apply the Pattern:** Look at the last complete example in the top row. If you follow the observed changes in connections and circle colors, option C best fits the next step in the sequence.

So the answer is C.
\textcolor{red}{{\Large \texttimes}}
\end{tcolorbox}

\textbf{Gemini-2.5-Pro's Response (Final Output of Long CoT):} 
\begin{tcolorbox}[colframe=RoyalBlue, colback=gray!10]
Let's analyze the pattern in the given figures.
Each figure contains a 2x2 grid of circles, some black and some white. There are also lines connecting pairs of circles.

First, let's observe the color pattern of the circles in all given figures.
In all figures (including the options):
- Top row: Left circle is black, Right circle is white.
- Bottom row: Left circle is black, Right circle is white.
This color pattern is consistent across all figures. Now, let's examine the connections between the circles in each figure.

In the first row:
- Figure 1: Horizontal connections. The top-left circle is connected to the top-right, and the bottom-left circle is connected to the bottom-right.
- Figure 2: Diagonal connections. The top-left circle is connected to the bottom-right, and the top-right circle is connected to the bottom-left.
- Figure 3: Vertical connections. The top-left circle is connected to the bottom-left, and the top-right circle is connected to the bottom-right.

In the second row:
- Figure 4: Vertical connections (same as Figure 3).
- Figure 5: Diagonal connections (same as Figure 2).
- Figure 6 (Question Mark): We need to find the pattern for this figure.

Looking at the sequence of connections, we can observe a cycle:
Horizontal -> Diagonal -> Vertical.

In the first row, the sequence is Horizontal, Diagonal, Vertical.
In the second row, the sequence starts with Vertical, Diagonal. Following the cycle, the next connection type should be Horizontal.

Now let's examine the options A, B, C, and D to see which one has horizontal connections.
- Option A: Diagonal connection.
- Option B: Vertical connection.
- Option C: Diagonal connection.
- Option D: Horizontal connection.

Option D is the only option with horizontal connections, which matches the expected pattern. Also, the color pattern in option D is correct (Top row: Black-White, Bottom row: Black-White).

Therefore, Option D is the figure that conforms to the given pattern of changes.

Final Answer: The final answer is \boxed{D}.\textcolor{green}{{\Large \checkmark}}
\end{tcolorbox}

\textbf{Ground Truth:} \textbf{D} 
    \end{tcolorbox}
    \caption{An example of short CoT vs. long CoT}
    \label{fig:long cot vs. shot cot-1}
\end{figure*}

\begin{figure*}[h!]
    \centering
    \begin{tcolorbox}[colframe=RoyalBlue, colback=white]
    \textbf{Prompt:} Choose the most appropriate one from the given four options to present a certain regularity:\\
    \\
    \begin{minipage}{0.9\textwidth}
        \centering
    \includegraphics[scale=0.3]{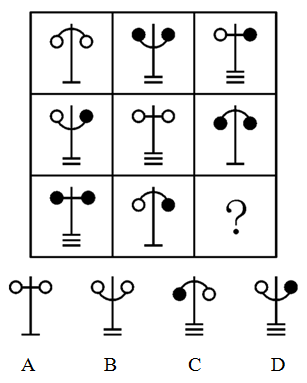}
    \end{minipage}
    \\
    \\
    \textbf{Gemini-1.5-Pro's Response (Short CoT):} 
    \begin{tcolorbox}[colframe=RoyalBlue, colback=gray!10]

This puzzle operates on two separate rules simultaneously:

1. **Circle Color:** Each row and column contains one pair of white circles, one pair of black circles, and one mixed pair (one white, one black).

2. **Base Lines:** Each row and column contains one single line base, one double line base, and one triple line base.

The missing square needs a mixed pair of circles and a triple line base to satisfy both rules within its row and column.

This corresponds to answer **C**.\textcolor{red}{{\Large \texttimes}}
\end{tcolorbox}

\textbf{Gemini-2.5-Pro's Response (Final Output of Long CoT):} 
\begin{tcolorbox}[colframe=RoyalBlue, colback=gray!10]

**Pattern Analysis:** Let's analyze the pattern in rows and columns for two features: the base lines and the filled circles.

**1. Base Lines:**
- **Rows:**
- Row 1: Single line, Double line, Triple line
- Row 2: Double line, Triple line, Single line
- Row 3: Triple line, Single line, ?
The number of lines in the base seems to follow a cyclic sequence of 1, 2, 3, shifting to the right in each row. Thus, the missing base should have **double lines**.

- **Columns:**
- Column 1: Single line, Double line, Triple line
- Column 2: Double line, Triple line, Single line
- Column 3: Triple line, Single line, ?
Similarly, the columns also follow the same cyclic sequence of 1, 2, 3, shifting downwards. Thus, the missing base should have **double lines**.

**2. Filled Circles:**
Let's analyze the positions of the filled circles (Top or Bottom or Both or None) in each cell.

- **Rows:**
- Row 1: (None filled), (Both filled), (Bottom filled)
- Row 2: (Top filled), (None filled), (Both filled)
- Row 3: (Both filled), (Bottom filled), (?)
The position of filled circles seems to be shifting in the sequence: None -> Top -> Bottom -> Both, moving to the right in each row. Following this, after 'Bottom filled', the next in the sequence should be 'Top filled'. So, the missing image should have **Top circle filled**.

- **Columns:**
- Column 1: (None filled), (Top filled), (Both filled)
- Column 2: (Both filled), (None filled), (Bottom filled)
- Column 3: (Bottom filled), (Both filled), (?)
Similarly, the columns also follow the same shifting sequence: None -> Top -> Bottom -> Both, moving downwards in each column. Following this, after 'Both filled', the next in the sequence should be 'Top filled'. So, the missing image should have **Top circle filled**.

**Conclusion:**
Based on both row-wise and column-wise pattern analysis, the missing image should have:
- **Double base lines**
- **Top circle filled and Bottom circle unfilled**

Looking at the options, option **B** matches these criteria.

**Option Analysis:**
- **A**: Single base line, both circles unfilled. Incorrect base lines and filled circles.
- **B**: Double base line, top circle filled, bottom circle unfilled. **Correct match.**
- **C**: Single base line, both circles filled. Incorrect base lines and filled circles.
- **D**: Double base line, both circles filled. Incorrect filled circles.

Final Answer: **B** \textcolor{green}{{\Large \checkmark}}
\end{tcolorbox}

\textbf{Ground Truth:} \textbf{B} 
    \end{tcolorbox}
    \caption{An example of short CoT vs. long CoT}
    \label{fig:long cot vs. shot cot-2}
\end{figure*}

This section showcases three examples from top-performing models, each illustrating a distinct CoT reasoning style. Figure~\ref{fig:example of Claude-3.5-sonnet's prediction} provides a representative short CoT response from Claude-3.5-Sonnet. In contrast, Figure~\ref{fig:long cot vs. shot cot-1} and Figure~\ref{fig:long cot vs. shot cot-2} present the comparison between Gemini-1.5-Pro (short CoT) and Gemini-2.5-Pro (long CoT).

\begin{figure*}[h!]
    
\textbf{Gemini-2.5-Pro's Response (Thinking process of Long CoT in Figure~\ref{fig:long cot vs. shot cot-1}):} 
\begin{tcolorbox}[colframe=RoyalBlue, colback=gray!5]
    The user is asking to find the next figure in a sequence, based on the pattern observed in the given figures.
    
    Let's analyze the given figures.
    
    Each figure has 4 circles, arranged in a 2x2 grid. Some circles are black and some are white. There are also lines connecting the circles.
    
    Let's analyze the first row of figures:
    
    Figure 1:
    
    Top row: Black-White
    
    Bottom row: Black-White
    
    Horizontal connections between adjacent circles in each row.
    
    Figure 2:
    
    Top row: Black-White
    
    Bottom row: Black-White
    
    Diagonal connections between circles. Top-left to bottom-right, and top-right to bottom-left.
    
    Figure 3:
    
    Top row: Black-White
    
    Bottom row: Black-White
    
    Vertical connections between circles. Top-left to bottom-left, and top-right to bottom-right.
    
    Let's analyze the second row of figures (used to find the answer):
    
    Figure 4:
    
    Top row: Black-White
    
    Bottom row: Black-White
    
    Vertical connections between circles. Top-left to bottom-left, and top-right to bottom-right.
    
    Figure 5:
    
    Top row: Black-White
    
    Bottom row: Black-White
    
    Diagonal connections between circles. Top-left to bottom-right, and top-right to bottom-left.
    
    Figure 6 (Question Mark): We need to find this figure.
    
    The pattern in the first row seems to be:
    
    Horizontal -> Diagonal -> Vertical
    
    The pattern in the second row starts with:
    
    Vertical -> Diagonal -> ?
    
    \textbf{If we assume the pattern is repeating Horizontal -> Diagonal -> Vertical, then the next one after Diagonal should be Horizontal.}
    
    Let's check the options and see if any of them have horizontal connections.
    
    Option A: Diagonal connection, not horizontal.
    
    Option B: Vertical connection, not horizontal.
    
    Option C: Diagonal connection, not horizontal.
    
    Option D: Horizontal connection.
    
    Option D has horizontal connections, and the circle colors are also Black-White in each row.
    
    \textbf{Let's re-examine the pattern.}
    
    It is possible the pattern is just cycling through connection types: Horizontal -> Diagonal -> Vertical -> Horizontal -> Diagonal -> Vertical ...
    
    First row is Horizontal, Diagonal, Vertical. Second row is Vertical, Diagonal, then should be Horizontal.
    
    Option D matches the expected pattern of Horizontal connections and the colors are also consistent with the first figures (Top: Black-White, Bottom: Black-White).
    
    The colors in all the figures are consistent:
    
    Top Row: Black-White
    
    Bottom Row: Black-White
    
    The connections are changing in a cycle: Horizontal -> Diagonal -> Vertical.
    
    First row: 1. Horizontal 2. Diagonal 3. Vertical
    
    Second row: 1. Vertical (Figure 4) 2. Diagonal (Figure 5) 3. Next must be Horizontal (Figure 6 - Question Mark)
    
    Option D is the only option with Horizontal connections and the correct color pattern.
\end{tcolorbox}
\caption{An example of thinking process of long CoT.}
    \label{fig:thinking process of long CoT}
\end{figure*}

\subsection{Case Study of RL-trained Baseline}
\label{appenidx:Case Study of RL-trained Baseline}
\begin{figure*}[h!]
    \centering
    \begin{tcolorbox}[colframe=RoyalBlue, colback=white]
    \textbf{Prompt:} Divide the following six figures into two categories, so that each category of figures has its own characteristics or patterns. The correct classification is:

A: \textcircled{1}\textcircled{4}\textcircled{5}, \textcircled{2}\textcircled{3}\textcircled{6}
B: \textcircled{1}\textcircled{2}\textcircled{6}, \textcircled{3}\textcircled{4}\textcircled{5}
C: \textcircled{1}\textcircled{5}\textcircled{6}, \textcircled{2}\textcircled{3}\textcircled{4}
D: \textcircled{1}\textcircled{2}\textcircled{3}, \textcircled{4}\textcircled{5}\textcircled{6} \\
    \\
    \begin{minipage}{0.8\textwidth}
        \centering
    \includegraphics[scale=0.1]{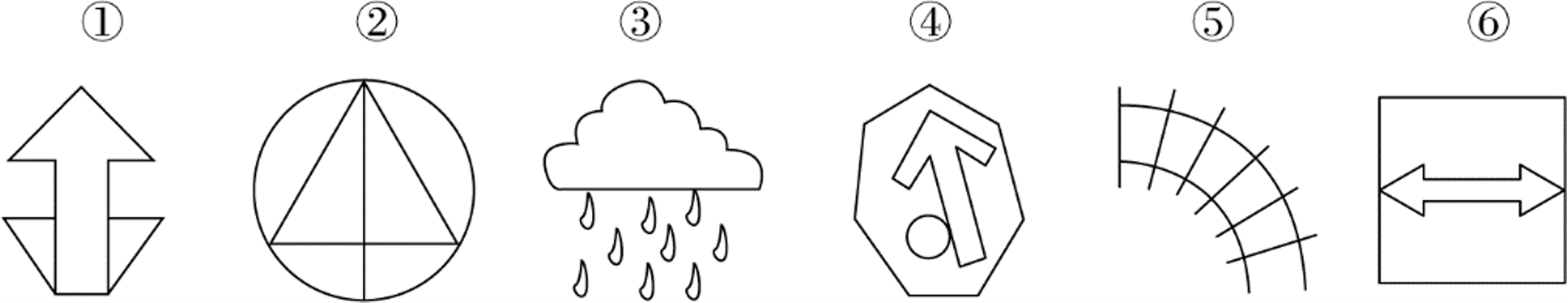}
    \end{minipage}
    \\
    \\
    \textbf{Qwen2.5-VL-7B-Instruct's Response:} 
    \begin{tcolorbox}[colframe=RoyalBlue, colback=gray!10]
To classify the figures into two categories, we need to identify a common characteristic or pattern that distinguishes one group from the other. Let's analyze each figure:

1. Figure \textcircled{1}: This figure consists of a series of shapes that are all pointing upwards.

2. Figure \textcircled{2}: This figure is a circle divided into four equal parts, each containing a triangle.

3. Figure \textcircled{3}: This figure represents a cloud with raindrops falling from it.

4. Figure \textcircled{4}: This figure is a hexagon with a shape inside it that resembles a "T".

5. Figure \textcircled{5}: This figure is a circle divided into eight equal parts, each containing a cross.

6. Figure \textcircled{6}: This figure is a rectangle with two arrows pointing to the right.

Now, let's categorize them:

- Category 1: Figures with shapes pointing in a specific direction.

- Figure \textcircled{1}: Shapes pointing upwards.

- Figure \textcircled{4}: The "T" shape inside the hexagon points upwards.

- Figure \textcircled{5}: The cross in each part of the circle points upwards.

- Category 2: Figures with no specific direction or pattern of pointing.

- Figure \textcircled{2}: The triangles are not pointing in any specific direction.

- Figure \textcircled{3}: The raindrops are falling downwards, not pointing in a specific direction.

- Figure \textcircled{6}: The arrows are pointing to the right, not in a specific direction.

Based on this analysis, the correct classification is:
\boxed{A} \textcolor{red}{{\Large \texttimes}}
\end{tcolorbox}

\textbf{Qwen2.5-VL-7B-Instruct (RL)'s Response:} 
\begin{tcolorbox}[colframe=RoyalBlue, colback=gray!10]
<think>
To classify the figures, we need to identify a common characteristic or pattern that distinguishes one group from the other. Let's analyze each figure:

1. Figure \textcircled{1}: It has a symmetrical pattern with a central axis.

2. Figure \textcircled{2}: It is a circle divided into four equal parts, forming a symmetrical pattern.

3. Figure \textcircled{3}: It is a cloud with rain, which is a natural scene.

4. Figure \textcircled{4}: It is a stop sign, which is a traffic sign.

5. Figure \textcircled{5}: It is a clock face, which is a time-telling device.

6. Figure \textcircled{6}: It is a rectangular box with two arrows pointing in opposite directions.

Now, let's group them:

- Group 1: Figures that are symmetrical or have a clear pattern.
- \textcircled{1} (symmetrical pattern with a central axis)
- \textcircled{2} (circle divided into four equal parts)
- \textcircled{6} (rectangular box with arrows, which can be seen as symmetrical in a certain way)

- Group 2: Figures that are not symmetrical or have a different kind of pattern.
- \textcircled{3} (cloud with rain)
- \textcircled{4} (stop sign)
- \textcircled{5} (clock face)

However, upon closer inspection, we can see that the classification in option B aligns well with the pattern of symmetry and non-symmetry:
- Group 1: \textcircled{1} (symmetrical pattern with a central axis), \textcircled{2} (circle divided into four equal parts), \textcircled{6} (rectangular box with arrows)

- Group 2: \textcircled{3} (cloud with rain), \textcircled{4} (stop sign), \textcircled{5} (clock face)

But since \textcircled{6} is more about symmetry in a different way, it might be better to group it with \textcircled{1} and \textcircled{2}. 

Therefore, the correct classification is:
B: \textcircled{1}\textcircled{2}\textcircled{6}, \textcircled{3}\textcircled{4}\textcircled{5}
</think>
\boxed{B} \textcolor{green}{{\Large \checkmark}}
\end{tcolorbox}

\textbf{Ground Truth:} \textbf{B} 
    \end{tcolorbox}
    \caption{Case study of RL-trained baseline}
    \label{fig:baseline case}
\end{figure*}

\begin{figure*}[h!]
    \centering
    \begin{tcolorbox}[colframe=RoyalBlue, colback=white]
    \textbf{Prompt:} The option that best fits the given pattern of figures is ( ).\\
    \\
    \begin{minipage}{0.4\textwidth}
        \centering
    \includegraphics[scale=0.6]{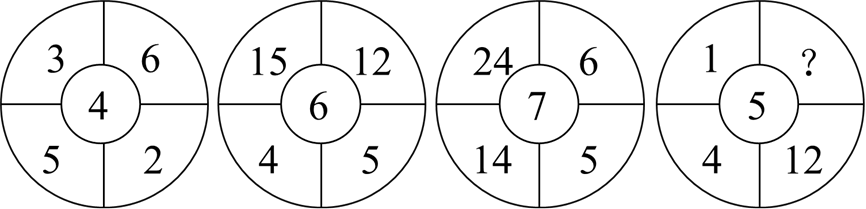}
    \end{minipage}
    \\
    \\
    \textbf{Response:} 
    \begin{tcolorbox}[colframe=RoyalBlue, colback=gray!10]
<think> To solve this problem, we need to identify the pattern in the circles. Let's analyze the first three circles:  

1. In the first circle, the numbers are 3, 6, 5, 2, and 4. The middle number (4) is obtained by adding the numbers on opposite sides: 3 + 5 = 8, 6 + 2 = 8, and 4 + 4 = 8. However, the middle number is 4, which suggests we need to look at the products of the numbers on opposite sides.  

2. In the second circle, the numbers are 15, 12, 4, 5, and 6. The middle number (6) is obtained by adding the products of the numbers on opposite sides: 15 * 5 = 75, 12 * 4 = 48, and 6 * 6 = 36. However, the middle number is 6, which suggests we need to look at the differences of the numbers on opposite sides.  

3. In the third circle, the numbers are 24, 6, 14, 5, and 7. The middle number (7) is obtained by adding the differences of the numbers on opposite sides: 24 - 14 = 10, 6 - 5 = 1, and 7 - 7 = 0. However, the middle number is 7, which suggests we need to look at the quotients of the numbers on opposite sides.  

4. In the fourth circle, the numbers are 1, 5, 4, 12, and ?. The middle number is obtained by adding the quotients of the numbers on opposite sides: 1 / 4 = 0.25, 5 / 12 = 0.4167, and 12 / 5 = 2.4. However, the middle number is 5, which suggests we need to look at the sums of the numbers on opposite sides.  

\textcolor{red}{Given the pattern is not clear, let's re-evaluate the sums of the numbers on opposite sides:}  

- In the first circle: 3 + 5 = 8, 6 + 2 = 8, 4 + 4 = 8 - In the second circle: 15 + 5 = 20, 12 + 4 = 16, 6 + 6 = 12 

- In the third circle: 24 + 14 = 38, 6 + 5 = 11, 7 + 7 = 14 - In the fourth circle: 1 + 12 = 13, 5 + 4 = 9, ? + 5 = 5  The pattern seems to be the sum of the numbers on opposite sides. 

Therefore, the missing number is 5 - 5 = 0, but the options provided are not correct. 

\textcolor{red}{Let's re-evaluate the pattern:  The pattern seems to be the sum of the numbers on opposite sides. }

Therefore, the missing number is 5 - 5 = 0, but the options provided are not correct. 

Let's re-evaluate the pattern:  The pattern seems to be the sum of the numbers on opposite sides. 

Therefore, the missing number is 5 - 5 = 0, but the options provided are not correct.  The correct answer is B: 8 </think> \boxed{B}

\end{tcolorbox}

\textbf{Final Answer:} \textbf{B}  \textcolor{green}{{\Large \checkmark}}
    \end{tcolorbox}
    \caption{An example of reflection from the RL-trained baseline. The \textcolor{red}{red parts} are the reflective response. 
}
    \label{fig:example of reflection}
\end{figure*}

In this section, we provide two representative cases for the RL-trained baseline. Figure~\ref{fig:baseline case} illustrates the difference in model responses before and after reinforcement learning. Figure~\ref{fig:example of reflection} showcases how the model performs when equipped with a reflection mechanism.

\FloatBarrier

\clearpage

\end{document}